%% file: main.tex
\documentclass[10pt,twocolumn,letterpaper]{article}

\usepackage[pagenumbers]{iccv} 

\input{preamble}

\definecolor{iccvblue}{rgb}{0.21,0.49,0.74}
\usepackage[pagebackref,breaklinks,colorlinks,allcolors=iccvblue]{hyperref}

\def\paperID{2734} 
\def\confName{ICCV}
\def\confYear{2025}

\title{Enhancing Reward Models for High-quality Image Generation: Beyond Text-Image Alignment}

\author{Ying Ba\textsuperscript{1,2,3\thanks{This work was done at iN2X.}}, Tianyu Zhang\textsuperscript{4}, Yalong Bai\textsuperscript{4}, Wenyi Mo\textsuperscript{1,2,3}, Tao Liang\textsuperscript{4}, Bing Su\textsuperscript{1,2,3\thanks{Corresponding author.}}, Ji-Rong Wen\textsuperscript{1,2,3}\\
\textsuperscript{1}Gaoling School of Artificial Intelligence Renmin University of China Beijing, China \\
\textsuperscript{2}Beijing Key Laboratory of Research on Large Models and Intelligent Governance \\
\textsuperscript{3}Engineering Research Center of Next-Generation Intelligent Search and Recommendation, MOE \\
\textsuperscript{4}iN2X \\
\tt\small \textsuperscript{1,2,3}\text{yingba88@outlook.com}, \text{\{mowenyi00, subingats\}@gmail.com}, \text{jrwen@ruc.edu.cn}\\ \tt\small \textsuperscript{4}\text{\{tianyu1949, liang0305tao\}@gmail.com}, \text{ylbai@outlook.com}
}

\begin{document}
\maketitle
\input{sec/0_abstract}
\vspace{-20pt}
\input{sec/1_intro}
\input{sec/2_relared}

\input{sec/3_theory}

\input{sec/4_method}

\input{sec/5_experiment}

\input{sec/6_conclusion}
{
    \small
    \bibliographystyle{ieeenat_fullname}
    \bibliography{main}
}
\appendix
\input{sec/8_appendix}

\end{document}

%% file: preamble.tex
\usepackage{booktabs}    
\usepackage{xcolor}      
\usepackage{colortbl}    
\usepackage{multirow}  
\usepackage{threeparttable}
\usepackage{graphicx}
\DeclareGraphicsExtensions{.png,.pdf,.jpg}
\graphicspath{{./images/}}
\usepackage{amsmath}    
\usepackage{amsthm}     
\usepackage{amssymb}    

\usepackage{dashrule}
\usepackage{arydshln}
\usepackage{siunitx}
\usepackage{array}
\usepackage{tabularx}
\usepackage{adjustbox}
\usepackage{lmodern}
\newtheorem{hypothesis}{Hypothesis}
\usepackage{tcolorbox}
\usepackage[accsupp]{axessibility}  


%% file: sec/0_abstract.tex
\begin{abstract}
Contemporary image generation systems have achieved high fidelity and superior aesthetic quality beyond basic text-image alignment. However, existing evaluation frameworks have failed to evolve in parallel. 
This study reveals that human preference reward models fine-tuned based on CLIP and BLIP architectures have inherent flaws: they inappropriately assign low scores to images with rich details and high aesthetic value, 
creating a significant discrepancy with actual human aesthetic preferences. 
To address this issue, we design a novel evaluation score, ICT (Image-Contained-Text) score, that achieves and surpasses the objectives of text-image alignment by assessing the degree to which images represent textual content. 
Building upon this foundation, we further train an HP (High-Preference) score model using solely the image modality to enhance image aesthetics and detail quality while maintaining text-image alignment. Experiments demonstrate that the proposed evaluation model improves scoring accuracy by over 10\% compared to existing methods, and achieves significant results in optimizing state-of-the-art text-to-image models. This research provides theoretical and empirical support for evolving image generation technology toward higher-order human aesthetic preferences. Code is available at \url{https://github.com/BarretBa/ICTHP}.
\end{abstract}

%% file: sec/1_intro.tex
\section{Introduction}
\label{sec:intro1}

In recent years, generation models have witnessed rapid development and found widespread applications across diverse domains~\cite{ddpm,ddim,rombach2022stablediffusion,ma2024star,podell2024sdxl,zhang2025v2flow,DBLP:conf/mm/Yang0W23,zhu2025llada15variancereducedpreference}. Modern high-performance diffusion models can not only accurately align texts and images but also generate visually impressive results. As the technology matures, the optimization focus of diffusion models has shifted from text-image alignment and prompt fidelity to generating high-quality images that better meet human aesthetic expectations~\cite{Mo2024DynamicPO,wu2023human,wu2023human2,hao2023optimizing,DBLP:conf/wacv/MoZB0W25}. In terms of optimization strategies, traditional data-driven training methods~\cite{oft,DreamBooth,Bai2024StyleInjectPE} are gradually being supplemented and expanded by reward model-based gradient backpropagation and reinforcement learning algorithms~\cite{zhang2024sdpo,ddpo}. Therefore, constructing a reward function that can precisely evaluate image quality is crucial for further improving the performance of diffusion models.

\begin{figure}
    \centering
    \includegraphics[width=1\linewidth]{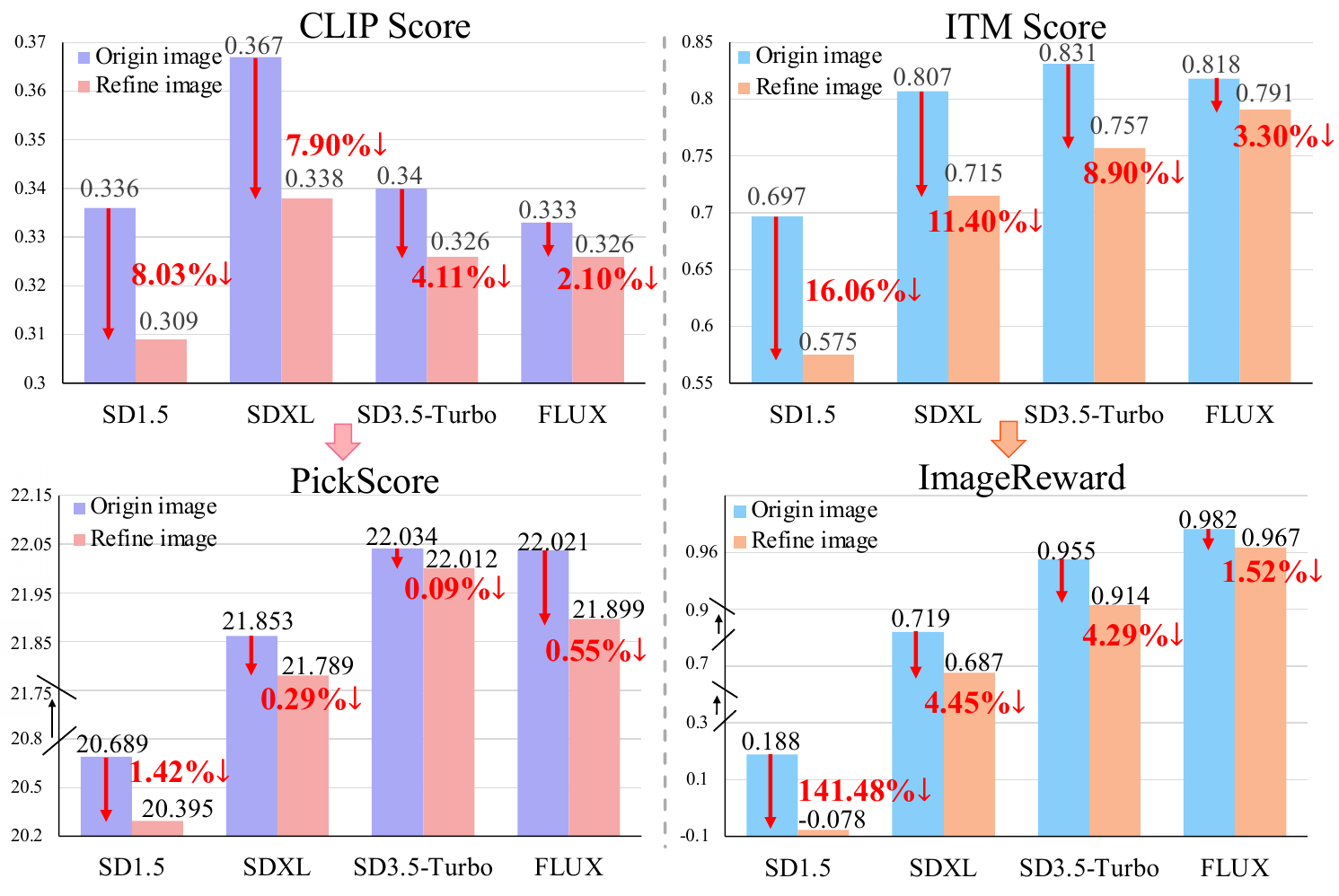}
    \caption{Reward Model Scoring Paradox Across Multiple Models Based on CLIP~\cite{clip} (CLIP Score, PickScore) and BLIP~\cite{li2022blipbootstrappinglanguageimagepretraining} (ITM Score, ImageReward). Refine images are generated with refined prompts by LLMs.}
    \label{fig:intro2}
    \vspace{-0.6cm}
\end{figure}

For text-to-image evaluation, text-image alignment (\textit{i.e.}, instruction compliance) serves as the fundamental requirement, while human preference determines the upper limit of image quality. Current reward models exhibit a critical flaw in measuring text-image alignment: they presume potential equivalence between the information content of images and text, whereas in reality, textual descriptions can hardly comprehensively capture all visual information in images. 
A direct manifestation is that for the same prompt, a basic image merely covering textual descriptions typically receives higher scores than a higher-quality image that contains richer details and visual elements extending beyond the literal description of the prompt.
This phenomenon persists not only in foundational models~\cite{li2022blipbootstrappinglanguageimagepretraining,clip} but also affects fine-tuned variants with human preferences like ImageReward~\cite{xu2023imagerewardlearningevaluatinghuman} and PickScore~\cite{kirstain2023pickapicopendatasetuser} (as shown in Fig.~\ref{fig:intro2}). 
Thus, previous reward models may effectively enhance Stable Diffusion-1.5/2.0~\cite{rombach2022stablediffusion} (where text-image alignment remains imperfect) but degrade state-of-the-art large models like Stable Diffusion-3.5~\cite{stabilityai2024sd35large}, steering them to generate visually sparse and aesthetically deficient images.

To address the above issue and continue leveraging human preference data to train superior reward models, we propose a novel optimization objective that transcends text-image alignment, \textit{i.e.}, the image-contained-text score (ICT score).
ICT score quantifies the extent to which an image contains textual information to evaluate the quality of generated images. 
By focusing on how well the image encapsulates textual semantics rather than enforcing bidirectional equivalence, ICT score prevents penalizing images that naturally contain richer contextual details beyond literal prompt descriptions, which aligns with human perception of high-quality outputs.
The ICT score removes the ceiling constraints of human preference models, enabling us to incorporate ultra-high-quality preference data to train an advanced human preference model, achieving a more comprehensive evaluation system.

The main contributions of this research can be summarized in the following three points:
\begin{itemize}
    \item We propose a novel image generation evaluation framework, including ICT score (quantifying the extent to which an image contains textual information) and HP score (evaluating how well an image meets human aesthetic preferences), fundamentally solving the problem of existing reward models underscoring high-quality images.
    \item A large-scale high-quality dataset (Pick-High) is constructed using chain-of-thought reasoning from large language models, providing a comprehensive benchmark for image generation quality assessment.
    \item We successfully apply the proposed reward model to optimize a state-of-the-art diffusion model (Stable Diffusion-3.5-turbo), demonstrating the effectiveness and superiority of this method in practical generation tasks.
\end{itemize}

%% file: sec/2_relared.tex
\section{Related Work}
\label{sec:related}
\noindent\textbf{Image quality evaluation for generative models.}
\quad The evaluation of generated images primarily falls into two categories: human assessment and automated metrics. While the former serves as the gold standard, its high cost and subjective biases have led to the widespread application of the latter in large-scale evaluations. Image-generation models have seen significant quality improvements, evolving from GAN-based approaches such as AttnGAN~\cite{Xu2017AttnGANFT} and DM-GAN~\cite{Zhu2019DMGANDM} to diffusion-based models including DALL·E 2~\cite{Ramesh2022HierarchicalTI}, Stable Diffusion~\cite{rombach2022stablediffusion,sd3}, Midjourney~\cite{midjourney2022}, and FLUX~\cite{flux2024}. Traditional metrics like FID~\cite{heusel2018ganstrainedtimescaleupdate} and IS~\cite{salimans2016improvedtechniquestraininggans} focus solely on distributional statistical properties and fail to capture human subjective preferences. Addressing this limitation, researchers have developed specialized human preference reward models trained on human preference data, such as PickScore~\cite{kirstain2023pickapicopendatasetuser}, HPSv2~\cite{wu2023humanpreferencescorev2}, and ImageReward~\cite{xu2023imagerewardlearningevaluatinghuman}. However, we have observed that these reward models exhibit significant biases when evaluating high-quality images generated through prompt optimization, indicating they have not yet effectively guided diffusion models in generating images that align with human preferences for detail and content richness.

\noindent\textbf{Image quality evaluation datasets.}
\quad Addressing the limitations of existing human preference models, high-quality evaluation datasets are essential for generative model assessment. As generative models rapidly evolved, researchers shifted toward constructing datasets based on human preferences, including the PickaPic\_v2 dataset~\cite{kirstain2023pickapicopendatasetuser}, ImageRewardDB~\cite{xu2023imagerewardlearningevaluatinghuman} developed by the ImageReward team, and the HPDv2 dataset~\cite{wu2023humanpreferencescorev2} employed by HPS. Despite their scale, these datasets predominantly contain lower-quality images inadequate for evaluating high-quality outputs from modern diffusion models and lack diversity across quality levels. To address these issues, we constructed a 600GB high-quality dataset called Pick-High, incorporating portions of the Pick-a-Pic dataset during training to ensure a comprehensive quality distribution.

%% file: sec/3_theory.tex
\section{Limitations of Text-image Alignment}
\label{sec:theory}

Multimodal models such as CLIP employ discriminative contrastive loss during pre-training to achieve text-image alignment. According to the information bottleneck theory, their encoders extract minimal image discriminative information to maximize mutual information with text. This training mechanism causes CLIP to exhibit bias when evaluating the same prompt: it tends to assign higher scores to simple images that merely faithfully reflect textual content, while giving lower scores to high-quality images rich in details, contradicting human aesthetic preferences.

According to the principle of information decomposition, the information content of a generated image can be divided into:
\begin{equation}
I(v) = I(v;t) + I(v|t),
\end{equation}

where $I(v;t)$ represents the mutual information between the image $v$ and text $t$(the image-text alignment component), and $I(v|t)$ represents the additional visual information contained in the image (aesthetics, texture, atmosphere, etc.).

\begin{figure}
    \centering
    \includegraphics[width=1\linewidth]{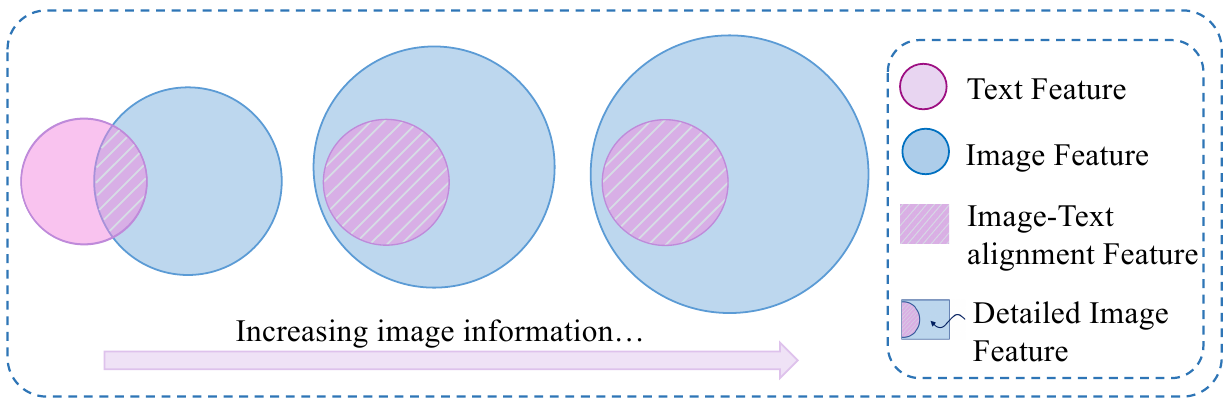}
    \caption{Text-Image Feature Interaction Venn Diagram: Increasing Image Information Under Fixed Text Prompts.}
    \label{fig:mutual}
    \vspace{-0.3cm}
\end{figure}

The CLIP scoring mechanism, based on cosine similarity, can be expressed as:
\begin{equation}\nonumber
\text{CLIP}(v,t) \approx \frac{I(v;t)}{\sqrt{I(t) \cdot I(v)}} =   
\frac{I(v;t)}{\sqrt{I(t) \cdot (I(v;t) + I(v|t))}}.
\end{equation}
As shown in Figure~\ref{fig:mutual}, as the capability of generative models improves, the information content of images increases. In the transition from image-text misalignment to alignment, $I(v;t)$ increases faster than $I(v)$, gradually raising CLIP scores. However, when high-quality models generate images rich in visual details, although $I(v;t)$ increases, $I(v|t)$ grows even faster, causing the denominator $\sqrt{I(v)}$ to increase more than the numerator $I(v;t)$, resulting in an overall decrease in scores, creating a contradiction with human preferences.
This motivates us to design a novel objective to train foundation models from the perspective of reward models, which will be elaborated in the next section.

%% file: sec/4_method.tex
\section{Method}
\vspace{-0.3cm}
This section presents our proposed methodological framework. In Section \ref{subsec:pick-high-dataset}, we introduce Pick-High dataset, a large-scale high-quality dataset constructed using refined prompts, which addresses the evaluation bias in existing reward models when assessing information-rich images. In Section \ref{subsec:Reward}, to resolve the inconsistency between the image-text alignment objective of foundation models and evaluation goals of reward models, we propose the Image-Contained-Text (\textbf{ICT}) optimization objective. This enhancement adapts the CLIP model for better evaluation of generative models. We further develop an innovative HP model that relies solely on image modality for human preference assessment. Finally, in Section \ref{subsec:diffusion}, we detail how we utilize the improved reward model to optimize diffusion models, thus enhancing the overall quality of generated images.
\begin{figure}
    \centering
    \includegraphics[width=1\linewidth]{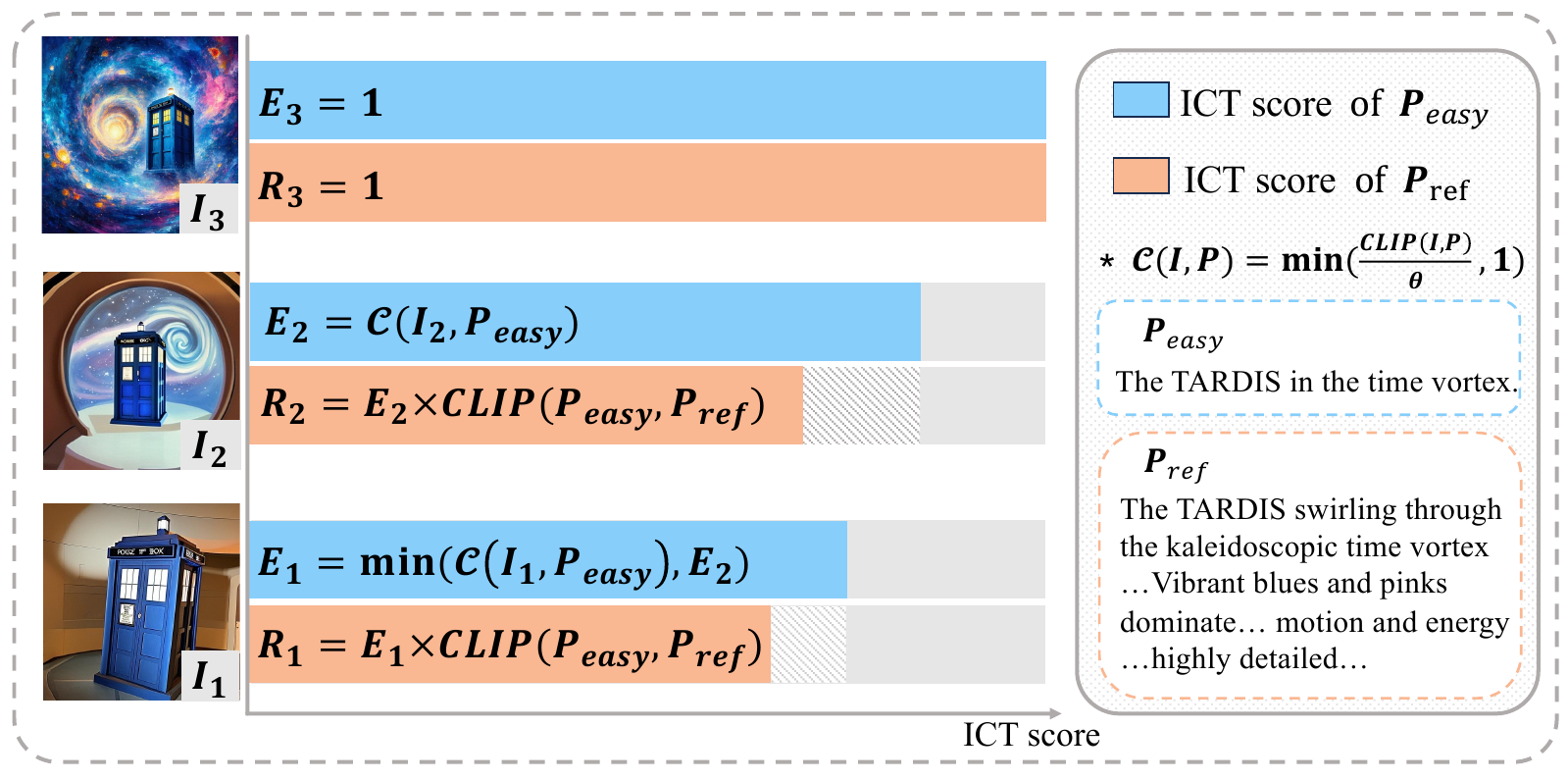}
    \caption{The Image-Contained-Text (ICT) Scoring Framework.}
    \label{fig:method1}
    \vspace{-0.8cm}
\end{figure}

\vspace{-0.2cm}
\subsection{Pick-High Dataset}\label{subsec:pick-high-dataset}
\vspace{-0.1cm}
Existing reward models (including PickScore, HPSv2, and ImgReward) inherit the limitations of foundational multimodal models, \textit{i.e.}, assigning unreasonably low scores when evaluating complex images paired with simple prompts, which significantly deviates from human preferences. To address this deficiency, we curated 360,000 prompts from the PickAPic\_v2 dataset and leveraged large language models' chain-of-thought capabilities to meticulously design a more refined set of prompts that closely align with human preferences. We input these optimized prompts into state-of-the-art image generation models to create 360,000 high-quality images, forming the Pick-High dataset. The detailed construction processes of Pick-High dataset can be seen in Appendix~\ref{sec:appendix}.

PickAPic\_v2 contains pairwise preferences of images. We denote the non-preferred low-quality image as $I_1$ and the preferred image as $I_2$. The Pick-High dataset extends pairwise preferences to triplet rankings by introducing a high-quality image $I_3$ generated with refined prompts. Based on these quality levels, we establish a hierarchical scoring mechanism for both basic prompts $P_\mathrm{easy}$ and refined prompts $P_\mathrm{ref}$, to better align human preferences.

\subsection{Reward Model Optimization Objective}\label{subsec:Reward}
\subsubsection{ICT Scores Design}

In text-to-image model evaluation, existing foundation models (e.g., CLIP, BLIP) employ instance-level image-text contrastive loss, which presents a fundamental issue: they tend to penalize high-quality images containing additional details. This evaluation mechanism significantly deviates from human preferences and reward model objectives.
Based on these observations, we propose the Image-Contained-Text (ICT) scoring mechanism to quantify the extent to which an image expresses textual content. 

The overview of our ICT scoring mechanism is shown in Fig.~\ref{fig:method1}.  We assign different scores to images ($I_1$, $I_2$, $I_3$) based on their alignment quality with prompts. For high-quality images $I_3$, we assign perfect scores for both basic prompts and refined prompts. To calculate ICT scores of $I_2$ and $I_1$ with basic prompts, we first develop a threshold-based mechanism instead of directly adopting CLIP similarity function: 
\begin{equation}
    \mathcal{C}(I,P)=\min (\frac{\mathrm{CLIP}(I,P)}{\theta},1),
\end{equation}
where $\mathrm{CLIP(I,P)}$ denotes the CLIP similarity function of an image $I$ and its corresponding prompt $P$ and $\theta$ represents the alignment threshold. This approach leverages the effectiveness of the CLIP model for assessment of low- and medium-quality images while circumventing the bias in CLIP scores of high-quality images. The basic prompt ICT scores ($E_1$, $E_2$, $E_3$) of $I_1$, $I_2$, $I_3$ containing $P_\mathrm{easy}$ are defined as follows:
\begin{align}
E_3 &= 1, \\
E_2 &= \mathcal{C}(I_2,P_\mathrm{easy}),\\
E_1 &= min(\mathcal{C}(I_1,P_\mathrm{easy}),E_2).
\end{align}

As for refined prompt ICT scores, based on basic prompt ICT scores, we further consider the CLIP similarities of $P_\mathrm{easy}$ and $P_\mathrm{ref}$, defining the scores of $I_2$ and $I_1$ containing $P_\mathrm{ref}$ as the product of $E_2$, $E_1$ and the textual similarity, as shown below:

\begin{align}
R_3 &= 1, \\
R_2 &= E_2 \times \mathrm{CLIP}(P_{\text{easy}}, P_{\text{ref}}), \\
R_1 &= E_1 \times \mathrm{CLIP}(P_{\text{easy}}, P_{\text{ref}}).
\end{align}

Unlike previous discriminative training objectives focused on achieving binary yes/no text-image alignment, our proposed ICT scores categorize the alignment quality between images and texts into distinct levels. This novel objective not only encourages models to achieve image-text alignment but also avoids penalizing images containing additional details beyond textual descriptions. It represents a superior training objective that goes beyond conventional text-image alignment metrics. In the following sections, we will elaborate on implementing these ICT scores as innovative labeling mechanisms for training multimodal models.

\subsubsection{ICT Model}

We finetune the CLIP model on Pick-High and Pickapic\_v2 datasets to obtain our ICT model. Each training sample consists of an image triplet with corresponding ICT labels $(E_1, E_2, E_3)$ for the basic prompt and $(R_1, R_2, R_3)$ for the refined prompt.  

The model computes ICT scores between each image and both prompts: $y_{i,e} = s(P_\mathrm{easy}, I_i)$ and $y_{i,r} = s(P_\mathrm{ref}, I_i)$ , where $i \in \{1,2,3\}$, and the scoring function $s$ is defined as $s(P, I) = \text{Enc}_{\text{txt}}(P) \cdot \text{Enc}_{\text{img}}(I) / \tau$, where $\tau$ is the temperature parameter. We optimize the model by minimizing the mean squared error between predicted scores and ICT scores:

\begin{equation}
\mathcal{L}_{\text{ICT}} = \sum_{i=1}^3(E_i - y_{i,e})^2 + \sum_{i=1}^3(R_i - y_{i,r})^2.
\end{equation}

To enhance the discriminative capability of the model, we introduce an in-batch negative sample learning strategy. For any combination of images and unpaired prompts within a batch, we set their ICT labels to 0 and compute the negative sample loss $\mathcal{L}_{\text{neg}} = \sum_{j \neq i} [(y_{j,e}^i)^2 + (y_{j,r}^i)^2]$ where $y_{j,e}^i$ and $y_{j,r}^i$ represent the ICT scores between the image of the i-th sample and the basic/refined prompts of the j-th sample, respectively. Considering the potential existence of false negative samples within batches, we adopt a smooth weighting strategy based on the sigmoid function for negative sample mining:

\begin{equation}
w(y) = \frac{1}{1 + e^{\alpha(|y| - \beta)}},
\end{equation}
where $\alpha$ is the smoothing coefficient controlling the steepness of the weight curve, $\beta$ is the threshold parameter, and $y$ is the predicted score of the sample. This function assigns lower weights to potential false negative samples with high predicted scores, thereby mitigating their negative impact on model training. The final negative sample loss function is: 

\begin{equation}
\mathcal{L}_{\text{neg}} = \sum_{j \neq i} (w(y_{j,e}^i)\cdot{y_{j,e}^i}^2 + w(y_{j,r}^i)\cdot{y_{j,r}^i}^2).
\end{equation}

The complete loss function is formulated as:
\begin{equation}
\mathcal{L}_{\text{total}} = \mathcal{L}_{\text{ICT}} + \lambda\mathcal{L}_{\text{neg}},
\end{equation}
where $\lambda$ is a balancing factor that controls the contribution of negative sample learning.

\subsubsection{HP Model}
When the image fully expresses the semantic content of the prompt, the ICT score reaches its upper limit. 
To further evaluate the quality of images, we focus solely on the image modality. 
For each preference triplet, we employ margin ranking loss to fine-tune both the CLIP model and its subsequent MLP layers. Specifically, given a triplet $\{I_1, I_2, I_3\}$, where $I_3$ is preferred over $I_2$, and $I_2$ is preferred over $I_1$, we enforce the preference margins as:
\begin{align}
\mathcal{L}_{\text{margin}} = \sum [ & \max(0, -\Delta(I_2, I_1) + m) + \nonumber \\
& \max(0, -\Delta(I_3, I_2) + m)],
\end{align}
where $\Delta(I_i, I_j)$ represents the score difference between images $I_i$ and $I_j$, $m$ is the margin threshold.

\subsection{Diffusion Model Optimization}\label{subsec:diffusion}

We adopt DRaFT-K~\cite{clark2024directlyfinetuningdiffusionmodels} as our fine-tuning method to optimize Stable-Diffusion-3.5-Large-Turbo~\cite{stabilityai2024sd35largeturbo} by directly maximizing the differentiable reward functions. 

Our proposed ICT model and HP model can each be individually utilized as reward functions. Additionally, we multiply the outputs of both models to create a more comprehensive integrated reward function, which we term the \textbf{ICT-HP model}.

%% file: sec/5_experiment.tex
\section{Experiments}
\begin{figure*}
    \centering
    \includegraphics[width=0.95\textwidth]{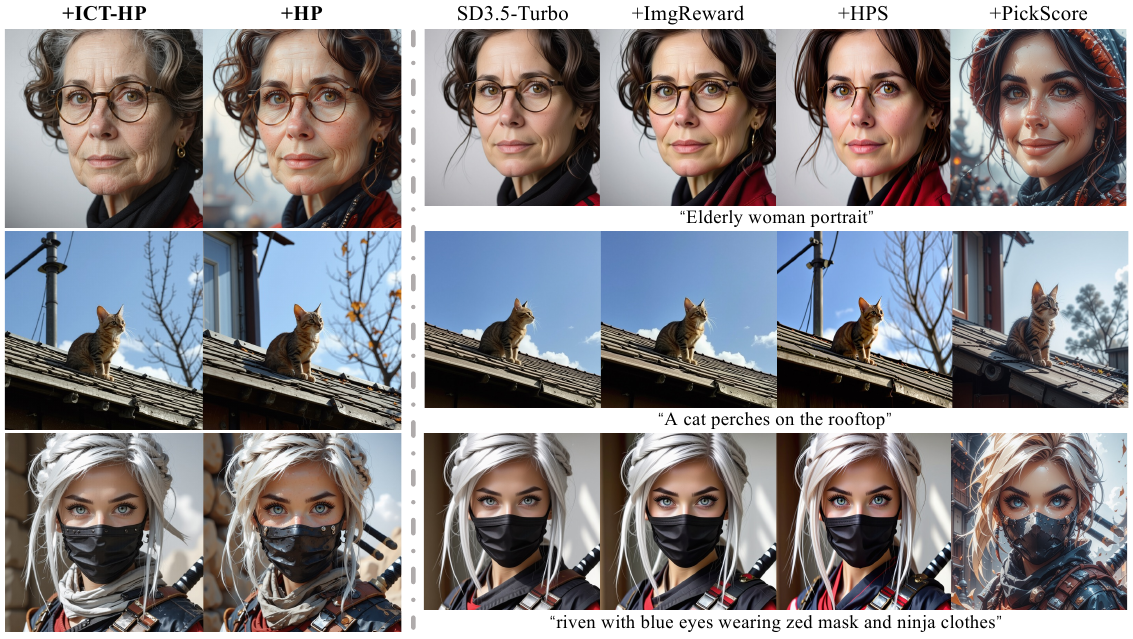}
    \caption{Qualitative Comparison of Optimization Results Across Reward Models Using Real User Prompts.}
    \label{fig:dx1}
    \vspace{-0.3cm}
\end{figure*}

\input{tabs/ictacc}
\input{tabs/jpgaes}

\subsection{Experimental Settings}

\noindent\textbf{Two-Stage Training Pipeline.} 
We adopt a two-stage training method for the ICT model and HP model. In the first stage, we fine-tune the CLIP model using our constructed ICT labels as training targets. In the second stage, we freeze the ICT model parameters and continue training the HP model. During the model optimization phase, we comparatively evaluate three different reward models for optimizing the diffusion model LoRA adapters: ICT model, HP model, and the more comprehensive ICT-HP model.

\noindent\textbf{Training Datasets.}
Our training dataset for the ICT model consists of the Pick-High dataset and a subset of Pickapic\_v2, comprising a total of 360,000 image triplets. For the diffusion model optimization process, we extract 32,000 non-repetitive text prompts from the Pick-High dataset to serve as our prompt dataset.

For detailed training hyperparameter configurations and experimental specifications, please refer to the supplementary materials.

\subsection{Reward Model Performance}
We evaluated the prediction accuracy of human preference models on the Pick-High and Pickapic\_v2 test sets. As shown in Table~\ref{tab:reward-comparison}, the image-text contrastive (ITC) classification heads of foundation multimodal pre-trained models CLIP and BLIP perform only comparable to random selection. This result indicates that the instance-level image-text alignment objective adopted by foundation multimodal models is not suitable for evaluating the quality of generative models.
In terms of overall average accuracy, our proposed ICT-HP model achieves nearly 10\% performance improvement compared to Pickscore. This significant improvement stems from the ability of our reward model to measure the extent to which images contain text, and on this basis, continue to measure image quality, thus more accurately identifying high-quality images and assigning them correspondingly high scores.

\noindent\textbf{Qualitative Results on SD3.5-Turbo Optimization.} In Figure~\ref{fig:dx1}, we present generation results using real user prompts. All images are generated with the same initial noise to ensure fair comparison. The experimental results show that: the HPS\_v2 model only enhances image tone with minimal detail improvement; the PickScore model, while increasing details, causes distortions in style and color fidelity; the ImageReward model shows almost no substantial improvement over the original model outputs.
In contrast, our proposed reward models achieve performance improvements in both textural details and tonal aesthetics. Specifically, the ICT model does not penalize increases in image information content, allowing diffusion models to continue optimizing toward richer visual content after achieving text-image alignment; meanwhile, the HP model effectively utilizes pure image features to further optimize when the ICT model reaches its performance ceiling. Compared to our approach, existing reward models tend to produce negative feedback on high-quality image features, thereby limiting their ability to enhance the performance of advanced diffusion models.

\noindent\textbf{Quantitative Evaluation of Visual Information Density and Aesthetic Fidelity.}
To quantitatively evaluate the enhancement effect of our reward model on the original model in terms of generated image information content and aesthetic quality, we employed two objective metrics, JPEG Compressibility~\cite{jpeg1992} and Aesthetic Score, for assessment on the Parti-Prompts dataset~\cite{parti}. As shown in Table ~\ref{tab:jpeg-aesthetics}, among all tested models, the HP model achieved the most significant improvements in both metrics, followed by the ICT-HP model. Notably, the SD3.5-Turbo optimized by the HP model not only demonstrated substantial improvement in aesthetic metrics, but even slightly surpassed more comprehensive baseline models such as SD3.5-Large and FLUX.1-dev.

\input{tabs/geneval}

\begin{figure}
    \centering
    \includegraphics[width=0.95\linewidth]{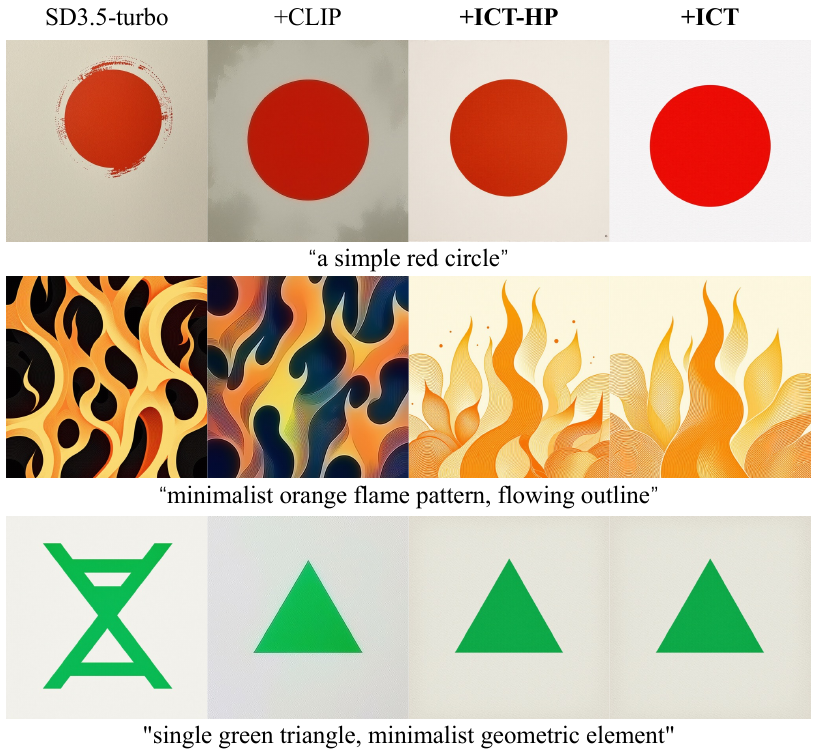}
    \caption{Qualitative Comparison of Simple Elements Generation Results.}
    \label{fig:simple1}
    \vspace{-0.2cm}
\end{figure}

\noindent\textbf{Qualitative Comparison of Simple Elements Generation Results.}
We conducted qualitative experiments on the generation of simple elements, with results shown in Figure~\ref{fig:simple1}.
The ICT model can stably handle simple geometric shapes and effectively penalize complex elements that do not meet simplicity requirements when prompts include descriptors like ``simple''.

\noindent\textbf{Comprehensive Evaluation on GenEval Benchmark.} Table~\ref{tab:geneval0} presents the quantitative performance of various image generation models within the multi-dimensional evaluation framework of GenEval~\cite{DBLP:journals/corr/abs-2310-11513}. Using SD3.5-Turbo as the baseline, our ICT-HP and ICT models both demonstrate significant advantages. The ICT-HP model excels in multi-object generation and color fidelity; while the ICT model achieves the highest mean score among all models and delivers optimal performance on the Counting metric, showcasing its superior capability in precise object quantity control. In contrast, the CLIP model suffers comprehensive performance degradation due to training collapse, and PickScore exhibits weaker performance in the Colors metric due to limitations in tonal representation. These comparisons validate our methods' significant advantages in terms of text-image consistency.

\begin{figure}
    \centering
    \includegraphics[width=0.92\linewidth]{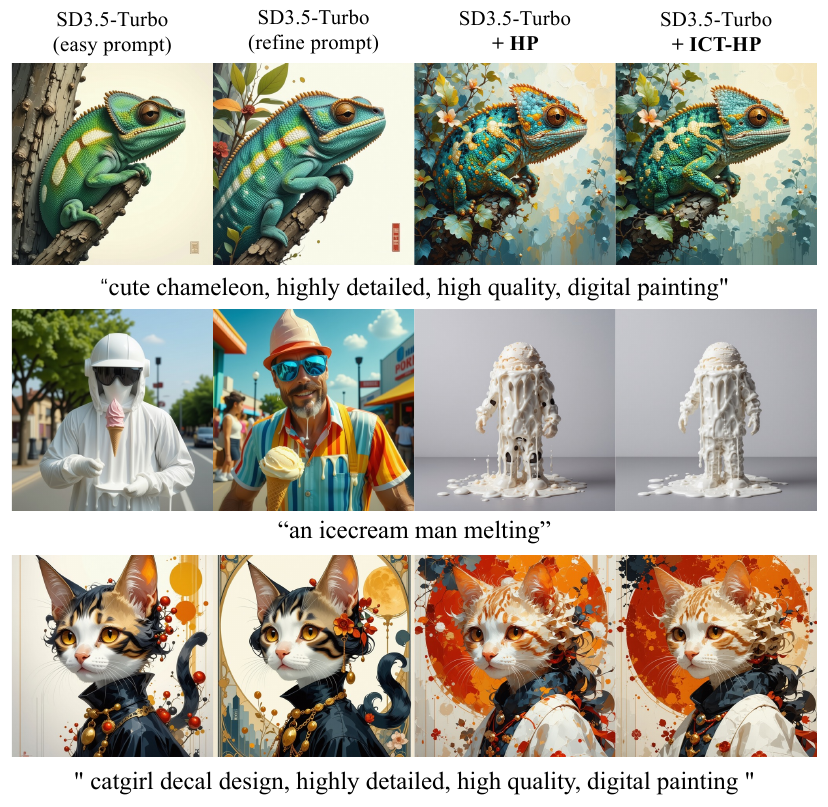}
    \caption{Qualitative Comparison: Original Images, Refined Images, and Generations Optimized by Our Reward Models (HP/ICT-HP).}
    \label{fig:newtt0}
    \vspace{-0.4cm}
\end{figure}

\begin{figure}
    \centering
    \includegraphics[width=1\linewidth]{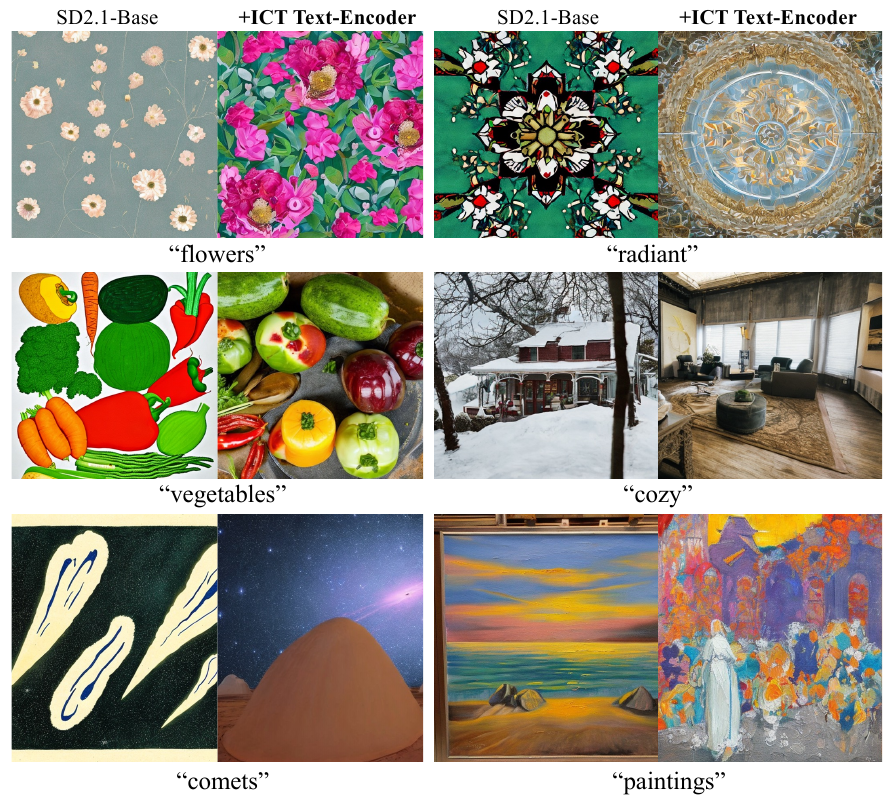}
    \caption{Qualitative Comparison of ICT Text-Encoder Direct Migration to SD2.1-Base Model.}
    \label{fig:sd21}
    \vspace{-0.3cm}
\end{figure}

\begin{figure}
    \centering
    \includegraphics[width=1\linewidth]{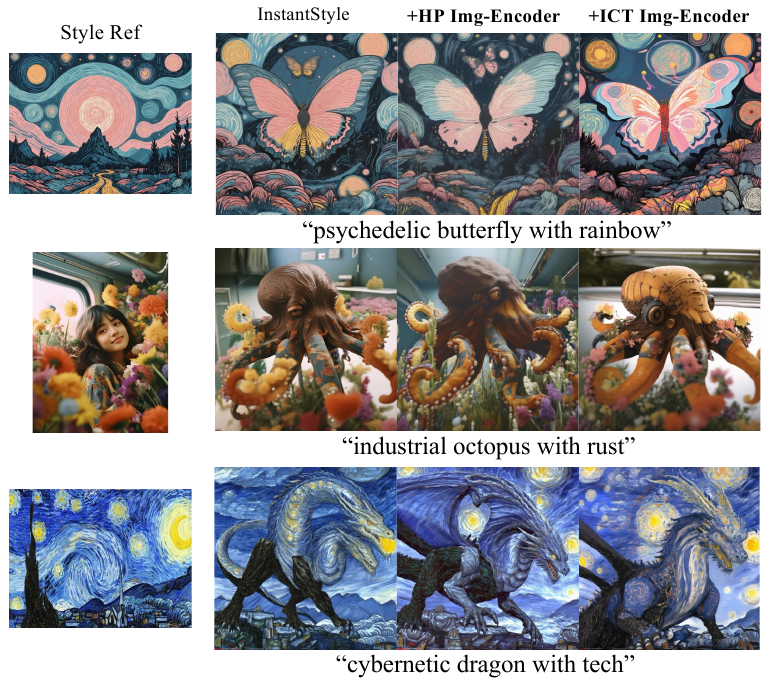}
    \caption{Qualitative Comparison of HP/ICT Image-Encoder Performance in Style Injection Tasks.}
    \label{fig:style0}
    \vspace{-0.4cm}
\end{figure}

\noindent\textbf{Quantitative and Qualitative Performance Comparison of Multi-metric Reward Models.}
\quad We conducted quantitative analyses of optimization results across various reward models, as shown in Table~\ref{tab:easyrefine-turbo}. Experiments reveal that existing reward models including HPS, PickScore, and ImageReward exhibit performance degradation across multiple metrics. In contrast, our ICT-HP and HP models achieved improvements across all metrics, demonstrating the stability advantage of our method. Although PickScore obtained the highest aesthetic score, qualitative analysis and GenEval experiments indicated significant distortion in artistic style and tone, with color-related metrics significantly lower than other models.
Figure~\ref{fig:newtt0} compares the original image, refined image, and our model-optimized results. ICT-HP and HP models not only generate intricate backgrounds that are difficult to describe directly through language but also demonstrate a more precise and comprehensive understanding of prompts, thus producing higher-quality images.

\input{tabs/ladder+}

\input{tabs/user1}

\noindent\textbf{Feature Transfer Validation.} As illustrated in Figure~\ref{fig:sd21}, transplanting the Text-Encoder trained by our ICT model into the SD2.1-base model significantly improves image quality. This demonstrates that our encoder integrates fine-grained visual details, thus transforming textual descriptions into more precise visual guidance signals.
Figure~\ref{fig:style0} presents results of style transfer with InstantStyle~\cite{wang2024instantstyle}. The CLIP image encoder is replaced by our ICT image encoder.
Our fine-tuned encoder not only precisely injects reference styles but also balances text descriptions, overcoming the style-content inconsistency.
These experiments validate the effectiveness and advancement of ICT score training objectives.

\subsection{User-Study}
We conduct a user study with ten annotators on 300 randomly selected prompts from Parti-Prompts~\cite{parti}. Image pairs (ours vs. baseline) were randomly ordered, and annotators evaluated prompt expression and visual appeal. As shown in Table \ref{tab:user1}, our methods achieve higher win rates against both the SD3.5-large-Turbo base model and the PickScore-optimized model, confirming our approach's effectiveness.

%% file: tabs/ictacc.tex
\begin{table}
\centering
\setlength{\tabcolsep}{9pt}
\small
\resizebox{\linewidth}{!}{%
\begin{tabular}{lcccc}
\toprule
\multirow{2}{*}{\textbf{Model}} & \multirow{2}{*}{\textbf{Average$\boldsymbol{\uparrow}$}} & \multicolumn{3}{c}{\textbf{Pairwise Comparison$\boldsymbol{\uparrow}$}} \\
\cmidrule(l){3-5}
& & \textbf{$I_2 > I_1$} & \textbf{$I_3 > I_2$} & \textbf{$I_3 > I_1$} \\
\midrule
Random & 50.00 & 50.00 & 50.00 & 50.00 \\
ICT-groundtruth & 87.55 & 62.80 & 99.87 & 99.98 \\
\midrule
CLIP~\cite{clip} & 60.30 & 64.29 & 52.80 & 63.79 \\
ITC~\cite{li2022blipbootstrappinglanguageimagepretraining}  & 57.36 & 63.33 & 48.42 & 60.31 \\
ITM~\cite{li2022blipbootstrappinglanguageimagepretraining}  & 63.39 & 63.88 & 58.28 & 67.99 \\
Aesthetic Score\footnotemark & 68.19 & 50.90 & \underline{77.81} & 75.83 \\
ImageReward~\cite{xu2023imagerewardlearningevaluatinghuman} & 63.81 & 64.58 & 58.02 & 68.84 \\
HPS\_v2~\cite{wu2023humanpreferencescorev2}  & 72.88 & \underline{66.98} & 71.63 & 80.04 \\
PickScore~\cite{kirstain2023pickapicopendatasetuser} & 79.04 & \textbf{74.80} & 75.37 & \underline{86.94} \\
\midrule
\textbf{ICT} & 87.58 & 64.65 & \textbf{100} & \textbf{100} \\
\rowcolor[HTML]{E6F3FF}
\textbf{HP} & \underline{88.47} & 64.97 & \textbf{100} & \textbf{100} \\
\rowcolor[HTML]{E6E6FA}
\textbf{ICT-HP} & \textbf{88.84} & 66.42 & \textbf{100} & \textbf{100} \\
\bottomrule
\end{tabular}%
}
\caption{\textbf{Preference prediction accuracy on test sets of Pick-High and Pickapic\_v2}. \textbf{Bold} and \underline{underlined} values represent optimal and second-best performance.}
\label{tab:reward-comparison}
\vspace{-0.5cm}
\end{table}
\footnotetext{https://github.com/christophschuhmann/improved-aesthetic-predictor}

%% file: tabs/jpgaes.tex
\begin{table}
\centering
\setlength{\tabcolsep}{9pt}
\small
\resizebox{\linewidth}{!}{\begin{tabular}{lccc}
\toprule
\textbf{Model} & \textbf{JPEG Compressibility$\boldsymbol{\uparrow}$} & \textbf{Aesthetic Score$\boldsymbol{\uparrow}$} \\
\midrule
SD3.5-Large~\cite{stabilityai2024sd35large} & 374.80 & 6.307 \\
FLUX.1-dev~\cite{flux2024} & 270.58 & 6.436 \\
\midrule
SD3.5-Turbo~\cite{stabilityai2024sd35largeturbo} & 313.10 & 6.293 \\
\multicolumn{1}{c}{} & \multicolumn{2}{c}{ - - - - - - - - - - - - - - - - - - - - - - - - - -} \\
+ CLIP~\cite{clip}(crash) & 279.98 & 5.175 \\
\textbf{+ ICT (Ours)} & 321.39 & 6.285 \\
\rowcolor[HTML]{E6F3FF}
\textbf{+ ICT-HP (Ours)} & \underline{330.23} & \underline{6.300} \\
\rowcolor[HTML]{E6E6FA}
\textbf{+ HP (Ours)} & \textbf{334.86} & \textbf{6.448} \\
\bottomrule
\end{tabular}}
\caption{\textbf{JPEG Compressibility and Aesthetic Scores on Parti-Prompts}. \textbf{Bold} and \underline{underlined} values represent optimal and second-best performance.}
\label{tab:jpeg-aesthetics}
\vspace{-0.5cm}
\end{table}

%% file: tabs/geneval.tex
\begin{table*}
\centering
\setlength{\tabcolsep}{9pt}
\small
\resizebox{\linewidth}{!}{\begin{tabular}{lccccccc}
\toprule
\textbf{Model} & {\bf Mean$\boldsymbol{\uparrow}$} & {\bf Single$\boldsymbol{\uparrow}$} & {\bf Two$\boldsymbol{\uparrow}$} & {\bf Counting$\boldsymbol{\uparrow}$} & {\bf Colors$\boldsymbol{\uparrow}$} & {\bf Position$\boldsymbol{\uparrow}$} & {\bf Color Attribution$\boldsymbol{\uparrow}$}\\
\midrule
SDXL~\cite{podell2023sdxlimprovinglatentdiffusion} & 0.55 & 0.98 & 0.74 & 0.39 & 0.85 & 0.15 & 0.23 \\
DALL-E 2~\cite{ramesh2022hierarchicaltextconditionalimagegeneration} & 0.52 & 0.94 & 0.66 & 0.49 & 0.77 & 0.10 & 0.19 \\
DALL-E 3~\cite{openai_dalle3_model_2023} & 0.67 & 0.96 & 0.87 & 0.47 & 0.83 & \textbf{0.43} & 0.45 \\
SD3~\cite{sd3} & 0.68 & 0.98 & 0.84 & 0.66 & 0.74 & \underline{0.40} & 0.43 \\
\midrule
SD3.5-Turbo~\cite{stabilityai2024sd35largeturbo}  & 0.69 & 0.99 & 0.87 & 0.69 & 0.80 & 0.25 & 0.55 \\
\multicolumn{1}{c}{} & \multicolumn{7}{c}{- - - - - - - - - -- - - - - - - - - - - - - - - - - - - - - - - - - - - - - - - - - - - - - - - - - - - - - - - - - - - - - - - - - -} \\
+ ImageReward & \underline{0.70} & \textbf{0.99} & \underline{0.87} & \underline{0.68} & \underline{0.80} & \underline{0.28} & \textbf{0.59} \\
+ HPS\_v2 & 0.69 & \underline{0.98} & 0.86 & \underline{0.68} & \textbf{0.81} & 0.27 & 0.57 \\
+ PickScore & 0.66 & \textbf{0.99} & 0.85 & 0.67 & 0.74 & 0.24 & 0.48 \\
\rowcolor[HTML]{E6F3FF}
\textbf{+ ICT-HP (Ours)} & \underline{0.70} & \textbf{0.99} & \textbf{0.88} & \underline{0.68} & 0.79 & \underline{0.28} & \underline{0.58} \\
\midrule
+ CLIP(crash) & 0.13 & 0.38 & 0.04 & 0.06 & 0.26 & 0.01 & 0.03 \\
\rowcolor[HTML]{E6E6FA}
\textbf{+ ICT (Ours)} & \textbf{0.71} & \underline{0.98} & \textbf{0.88} & \textbf{0.70} & \textbf{0.81} & \textbf{0.31} &  0.56 \\
\bottomrule
\end{tabular}}
\caption{\textbf{Quantitative Evaluation of Optimization Results Across Reward Models on GenEval Benchmark.} \textbf{Bold} and \underline{underlined} values represent optimal and second-best performance.}
\label{tab:geneval0}

\end{table*}

%% file: tabs/ladder+.tex
\begin{table*}
\centering
\setlength{\tabcolsep}{9pt}
\small
\resizebox{\linewidth}{!}{%
\begin{tabular}{l*{4}{S[table-format=1.3]}S[table-format=2.3]*{4}{S[table-format=1.3]}}
\toprule
{\textbf{Model}} & {\textbf{CLIP}$\boldsymbol{\uparrow}$} & {\textbf{ITM}$\boldsymbol{\uparrow}$} & {\textbf{ImgRwd}$\boldsymbol{\uparrow}$} & {\textbf{HPS}$\boldsymbol{\uparrow}$} & {\textbf{Pick}$\boldsymbol{\uparrow}$} & {\textbf{Aes}$\boldsymbol{\uparrow}$} & {\textbf{ICT}$\boldsymbol{\uparrow}$} & {\textbf{HP}$\boldsymbol{\uparrow}$} & {\textbf{ICT-HP}$\boldsymbol{\uparrow}$} \\
\midrule
SD3.5-T$_{e}$ & 0.340 & 0.831 & 0.955 & 0.277 & 22.034 & 6.555 & 0.910 & 0.777 & 0.717 \\
SD3.5-T$_{r}$ & {\cellcolor{pink!15}0.326 \textcolor{red}{$\boldsymbol{\downarrow}$}} & {\cellcolor{pink!15}0.757 \textcolor{red}{$\boldsymbol{\downarrow}$}} & {\cellcolor{pink!15}0.914 \textcolor{red}{$\boldsymbol{\downarrow}$}} & {\cellcolor{pink!15}0.276 \textcolor{red}{$\boldsymbol{\downarrow}$}} & {\cellcolor{pink!15}22.012 \textcolor{red}{$\boldsymbol{\downarrow}$}} & {\underline{6.780} \textcolor{green}{$\boldsymbol{\uparrow}$}} & {0.923 \textcolor{green}{$\boldsymbol{\uparrow}$}} & {\textbf{0.781} \textcolor{green}{$\boldsymbol{\uparrow}$}} & {0.718 \textcolor{green}{$\boldsymbol{\uparrow}$}} \\
\midrule
{+HPS\_v2} & {0.343 \textcolor{green}{$\boldsymbol{\uparrow}$}} & {0.835 \textcolor{green}{$\boldsymbol{\uparrow}$}} & {1.065 \textcolor{green}{$\boldsymbol{\uparrow}$}} & {\textbf{0.282} \textcolor{green}{$\boldsymbol{\uparrow}$}} & {\cellcolor{pink!15}22.000 \textcolor{red}{$\boldsymbol{\downarrow}$}} & {6.590 \textcolor{green}{$\boldsymbol{\uparrow}$}} & {0.929 \textcolor{green}{$\boldsymbol{\uparrow}$}} & {0.777 \textcolor{black}{$\boldsymbol{-}$}} & {0.722 \textcolor{green}{$\boldsymbol{\uparrow}$}} \\
{+PickScore} & {\cellcolor{pink!15}0.332 \textcolor{red}{$\boldsymbol{\downarrow}$}} & {\cellcolor{pink!15}0.816 \textcolor{red}{$\boldsymbol{\downarrow}$}} & {\textbf{1.106} \textcolor{green}{$\boldsymbol{\uparrow}$}} & {0.277 \textcolor{black}{$\boldsymbol{-}$}} & {\textbf{22.719} \textcolor{green}{$\boldsymbol{\uparrow}$}} & {\textbf{6.924} \textcolor{green}{$\boldsymbol{\uparrow}$}} & {0.929 \textcolor{green}{$\boldsymbol{\uparrow}$}} & {0.779 \textcolor{green}{$\boldsymbol{\uparrow}$}} & {0.723 \textcolor{green}{$\boldsymbol{\uparrow}$}} \\
{+ImageReward} & {0.340 \textcolor{black}{$\boldsymbol{-}$}} & {\textbf{0.841} \textcolor{green}{$\boldsymbol{\uparrow}$}} & {1.056 \textcolor{green}{$\boldsymbol{\uparrow}$}} & {\underline{0.279} \textcolor{green}{$\boldsymbol{\uparrow}$}} & {\cellcolor{pink!15}22.022 \textcolor{red}{$\boldsymbol{\downarrow}$}} & {6.569 \textcolor{green}{$\boldsymbol{\uparrow}$}} & {\cellcolor{pink!15}0.865 \textcolor{red}{$\boldsymbol{\downarrow}$}} & {\cellcolor{pink!15}0.763 \textcolor{red}{$\boldsymbol{\downarrow}$}} & {\cellcolor{pink!15}0.663 \textcolor{red}{$\boldsymbol{\downarrow}$}} \\
{+\textbf{ICT(Ours)}} & {\underline{0.344} \textcolor{green}{$\boldsymbol{\uparrow}$}} & {\textbf{0.841} \textcolor{green}{$\boldsymbol{\uparrow}$}} & {1.014 \textcolor{green}{$\boldsymbol{\uparrow}$}} & {0.277 \textcolor{black}{$\boldsymbol{-}$}} & {22.117 \textcolor{green}{$\boldsymbol{\uparrow}$}} & {6.561 \textcolor{green}{$\boldsymbol{\uparrow}$}} & {0.937 \textcolor{green}{$\boldsymbol{\uparrow}$}} & {\cellcolor{pink!15}0.776 \textcolor{red}{$\boldsymbol{\downarrow}$}} & {\underline{0.728} \textcolor{green}{$\boldsymbol{\uparrow}$}} \\
\rowcolor[HTML]{E6F3FF}
{+\textbf{ICT-HP(Ours)}} & {\textbf{0.345} \textcolor{green}{$\boldsymbol{\uparrow}$}} & {\underline{0.839} \textcolor{green}{$\boldsymbol{\uparrow}$}} & {1.011 \textcolor{green}{$\boldsymbol{\uparrow}$}} & {0.277 \textcolor{black}{$\boldsymbol{-}$}} & {\underline{22.147} \textcolor{green}{$\boldsymbol{\uparrow}$}} & {6.558 \textcolor{green}{$\boldsymbol{\uparrow}$}} & {\textbf{0.938} \textcolor{green}{$\boldsymbol{\uparrow}$}} & {0.778 \textcolor{green}{$\boldsymbol{\uparrow}$}} & {\textbf{0.730} \textcolor{green}{$\boldsymbol{\uparrow}$}} \\
\rowcolor[HTML]{E6E6FA}
{+\textbf{HP(Ours)}} & {0.342 \textcolor{green}{$\boldsymbol{\uparrow}$}} & {0.836 \textcolor{green}{$\boldsymbol{\uparrow}$}} & {\underline{1.069} \textcolor{green}{$\boldsymbol{\uparrow}$}} & {0.278 \textcolor{green}{$\boldsymbol{\uparrow}$}} & {22.231 \textcolor{green}{$\boldsymbol{\uparrow}$}} & {\underline{6.668} \textcolor{green}{$\boldsymbol{\uparrow}$}} & {0.937 \textcolor{green}{$\boldsymbol{\uparrow}$}} & {0.779 \textcolor{green}{$\boldsymbol{\uparrow}$}} & {\textbf{0.730} \textcolor{green}{$\boldsymbol{\uparrow}$}} \\
\bottomrule
\end{tabular}%
}
\caption{\textbf{Quantitative Results}. SD3.5-T$_{e}$/T$_{r}$ denote images generated using easy/refine prompts on SD-3.5-Turbo. Arrows (\textcolor{green}{$\boldsymbol{\uparrow}$}, \textcolor{red}{$\boldsymbol{\downarrow}$}, \textcolor{black}{$\boldsymbol{-}$}) indicate changes vs. SD3.5-T$_{e}$. \textbf{Bold} and \underline{underlined} values represent optimal and second-best performance.}
\label{tab:easyrefine-turbo}
\end{table*}

%% file: tabs/user1.tex
\begin{table}[htbp]
    \centering
    \begin{tabular}{l c c}
        \toprule
        \textbf{Comparison} & \textbf{Models} & \textbf{Win Rate} \\
        \midrule
        \multirow{3}{*}{vs. Base} & ICT model  & 0.562 \\
        & HP model & \textbf{0.709} \\
        & ICT-HP model & 0.673 \\
        \midrule
        \multirow{2}{*}{vs. PickScore} & HP model  & 0.617 \\
        & ICT-HP model & \textbf{0.714} \\
        \bottomrule
    \end{tabular}
    \caption{Human evaluation on 300 randomly selected samples from Parti-Prompts~\cite{parti}. }
    \label{tab:user1}
    \vspace{-0.8cm}
\end{table}

%% file: sec/6_conclusion.tex
\section{Conclusion}
In this paper, we reveal the reward-scoring paradox of high-quality images and show that the inherent flaw in current text-to-image generation reward models lies in classic text-image alignment paradigms. 
We propose a novel training objective for reward models, \textit{i.e.}, ICT (Image-Contained-Text) score. 
This objective encourages models to achieve text-image alignment and avoid penalizing high-quality images containing extra details, that better match human preferences.
Based on the ICT model, we further train a preference model that relies solely on image modality for human preference assessment.
Our experiments show that the proposed approach improves scoring accuracy by over 10\% compared to existing methods and markedly enhances state-of-the-art models like SD3.5-Turbo. Through both theoretical foundation and empirical validation, our work advances image generation technology toward better alignment with human preferences.

\section*{Acknowledgments}
This work was supported in part by the National Natural Science Foundation of China No. 62376277, the Public Computing Cloud of Renmin University of China and the Fund for Building World-Class Universities (Disciplines) of Renmin University of China.

%% file: sec/8_appendix.tex
\clearpage
\appendix
\section{Appendix}
\label{sec:appendix}

\subsection{Construction of Pick-High Dataset}
\subsubsection{Refining Prompt Construction}
\begin{figure*}[t]
    \vskip -0.1in
    \centering
    \includegraphics[width=0.9\linewidth]{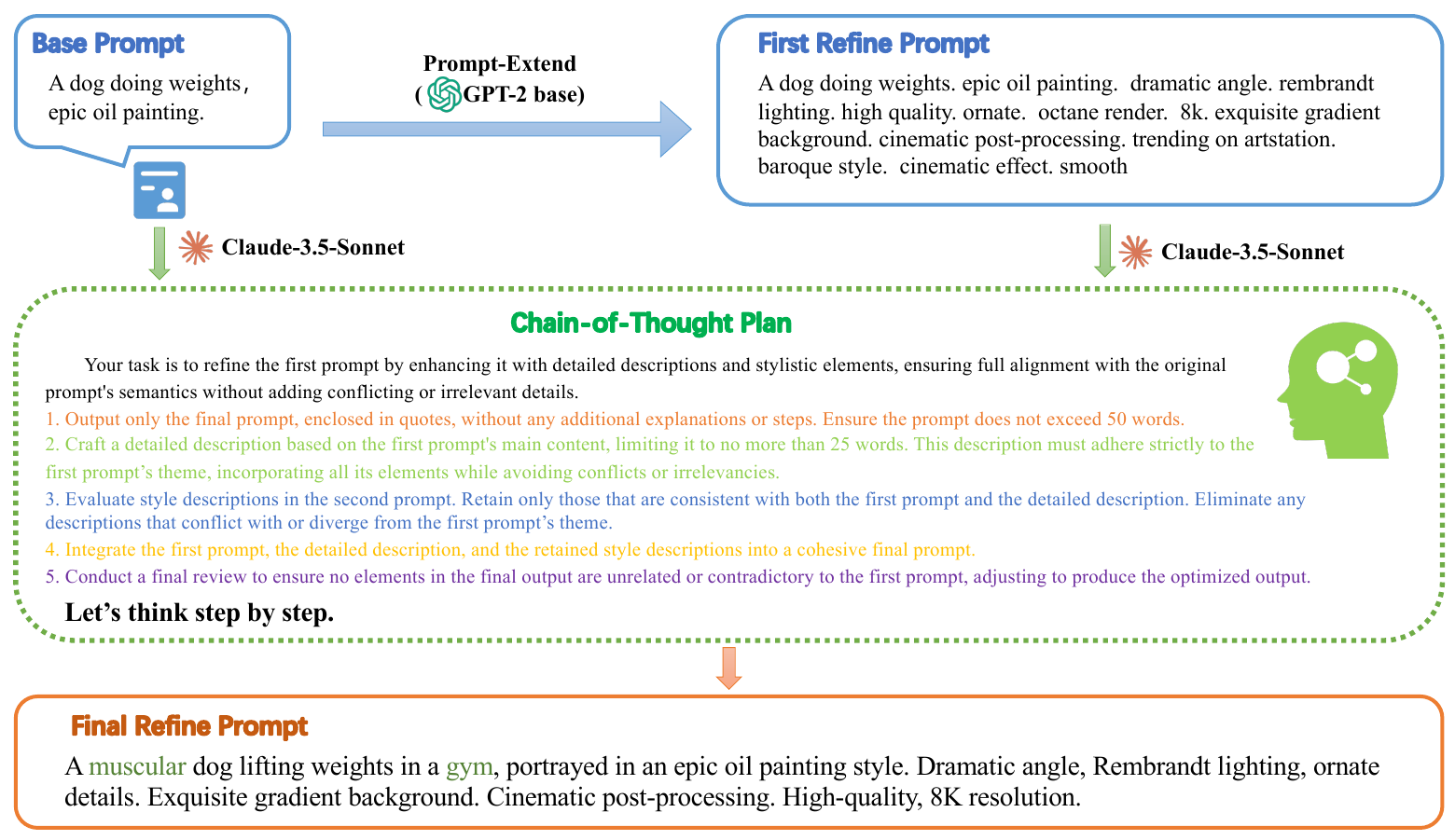}
    \caption{The Overview of Prompt Refinement Pipeline via Large Language Model Chain-of-Thought.}
    \label{fig:cot}
\end{figure*}

We selected 360,000 relatively short prompts from the pickapic\_v2 dataset's prompt collection as our base prompts. As shown in Figure~\ref{fig:cot}, to refine these base prompts, we first employed a GPT-2 model (Prompt-Extend) trained on diffusion model prompts, which generates appropriate style and detail descriptions based on the prompt's main theme.
To further enhance the prompts' coherence and granularity, we designed a chain-of-thought template and utilized the Claude-3.5-sonnet for chain-of-thought reasoning. Through this process, we filtered out descriptions that didn't match the base prompt's style and any inadvertently introduced anomalous content, while simultaneously enriching the prompts with thematically appropriate aesthetic and stylistic information.

\subsubsection{Image Collection and Dataset construction}
Given the limitations in generalization performance of vision language models\cite{song2024learning,zhangrethinking,qiang2025out}, we additionally verified whether the refined prompts fully preserve the core concepts of the base prompts while introducing no conflicting information.. Our verification included: (1) Filtering out NSFW contents rejected by Claude-3.5-sonnet ($\sim$2\%); (2) Applying a binary verification CoT template to exclude non-compliant samples ($\sim$7\%); Additionally, expert evaluation of 1000 randomly sampled Pick-High items confirmed 97\% prompt and 95\% image compliance with requirements. Based on this, we input all refined prompts into the Stable Diffusion-3.5-large model to generate 360,000 high-quality images, forming our proposed Pick-High dataset. Since the base prompts in the Pick-High dataset originate from the filtered results of the pickapic\_v2 dataset, the natural fusion of these two datasets creates a training dataset with ternary preference relationships and significant quality variations.

\subsection{Experiment Details}

Our experimental framework comprises three sequential phases: ICT model training, HP model training, and diffusion model optimization.

\noindent\textbf{ICT Model Training}
In the first phase, we fine-tune all parameters of the CLIP-H model~\cite{Cherti_2023} using MSE loss to optimize ICT scores. Training is conducted on 8 NVIDIA A800 GPUs for a total of 40,000 iterations. We employ the AdamW optimizer~\cite{loshchilov2019decoupledweightdecayregularization} with a learning rate of 3e-5. The smoothing coefficient $\alpha$ and threshold parameter $\beta$ in the negative sample smoothing function are set to 20 and 6, respectively, while the balancing factor $\lambda$ in the loss function is 0.1.
\noindent\textbf{HP Model Training}
In the second phase, we keep the ICT model parameters fixed and only fine-tune the latter half of the CLIP-H model~\cite{Cherti_2023} parameters and its connected MLP layers. Training is similarly performed on 8 NVIDIA A800 GPUs for 50,000 iterations. The margin threshold $m$ is set to 0.2, using the AdamW optimizer~\cite{loshchilov2019decoupledweightdecayregularization} with a learning rate of 3e-6.

\noindent\textbf{Diffusion Model Optimization}
In the final phase, we optimize the diffusion model using the trained reward models. We select Stable Diffusion-3.5-turbo as the base model and conduct training in half-precision (FP16). The diffusion process is configured with 8 sampling steps and a Guidance Scale of 0.0. Following the DRaFT-K method~\cite{clark2024directlyfinetuningdiffusionmodels}, we only propagate gradients through the last 3 denoising steps to optimize the LoRA parameters in the transformer layers, while keeping all other base model parameters frozen. Training is performed on 5 nodes equipped with 8 NVIDIA A800 GPUs each, for a total of 3,000 iterations, with a training time of approximately 24 hours. We utilize the AdamW optimizer~\cite{loshchilov2019decoupledweightdecayregularization} with a learning rate of 5e-6.

\input{tabs/ladder}

\subsection{Comprehensive Reward Scoring for Original and Refine Images Across Diffusion Models}
We randomly selected 800 base prompts from DiffusionDB~\cite{wang2023diffusiondblargescalepromptgallery} and COCO Captions~\cite{ding2022cogview2fasterbettertexttoimage}, and obtained refined prompts through optimization by large language models. These two sets of prompts were input into six diffusion models with diverse architectures (SD1.5, SDXL, SD3.5-Turbo, SD3.5-Large-Turbo, FLUX.1-schnell, and FLUX-1.dev) for image generation. We use the suffix ``e'' to denote images generated with base prompts and ``r'' for images generated with refined prompts.

As shown in Table~\ref{tab:ladder}, when evaluating images generated from refined prompts, scores decrease across both basic multimodal models (CLIP and BLIP) and all human preference models. This indicates that images generated from refined prompts contain richer information, resulting in reduced text-image similarity. Notably, our refined prompts, filtered through the chain-of-thought process of large language models, do not contain semantically irrelevant subjects or style words. Therefore, refine images do not introduce text-image misalignment, but rather enhance aesthetic qualities, details, and texture-related information.

Experimental results demonstrate that aesthetic metrics based solely on the image modality show improvement across all test cases, confirming that images generated from refined prompts indeed contain richer visual information. Since pure image modality evaluation is not affected by explicit or implicit text-image comparison objectives, its assessment results align with human preference trends.

Our trained ICT model, HP model, and the combined ICT-HP model show score improvements across almost all tested models, strongly demonstrating that the ICT training objective successfully addresses the inherent deficiency between instance-level text-image alignment and human preferences. The results indicate that as the generation model quality reaches high levels, the ICT metric maintains stable high values, suggesting our ICT scoring mechanism does not negatively evaluate high-quality images after reaching the text-image alignment inflection point. The HP model, as a reward model trained solely on the image modality, ensures higher scores for refine images in all test scenarios. In the current implementation, the ICT-HP score is calculated through a simple product of ICT and HP scores; future research could explore more optimal methods for integrating these two models.

\input{tabs/geneval_fl}

\subsection{Optimizing FLUX.1-schnell with HP and ICT-HP Models: Implementation and Evaluation}
\vspace{-0.1cm}
We applied our proposed HP model and ICT-HP model to optimize the flux.1-schnell architecture. During training, we employed half-precision (FP16) to enhance computational efficiency. The diffusion process was configured with a 4-step sampling procedure and a Guidance Scale of 0.0. Following the DRaFT-K methodology ~\cite{clark2024directlyfinetuningdiffusionmodels}, we selectively propagated gradients through only the final denoising step to optimize the LoRA parameters within the transformer layers, while maintaining the integrity of other parameters in the base model. The experimental setup consisted of 4 computing nodes, each equipped with 8 NVIDIA A800 GPUs, supporting a total of 3,000 training iterations completed in approximately 24 hours. For optimization, we utilized the AdamW optimizer \cite{loshchilov2019decoupledweightdecayregularization} with a learning rate of 5e-6. Herein, we present both qualitative analysis and quantitative evaluation results on the GenEval benchmark.

Table~\ref{tab:geneval_fl} presents the quantitative results of our FLUX.1-schnell optimization using both the ICT-HP model and ICT model. Additionally, we directly transferred the LoRA weights trained on FLUX.1-schnell to FLUX.1-dev to obtain quantitative performance metrics within the multidimensional GenEval evaluation framework. Our ICT-HP and ICT models demonstrated notable advantages compared to the baseline models, with particularly significant improvements in color-related scores. These results indicate that our reward model effectively enhances the color fidelity of the FLUX model series.

\subsection{Comprehensive Presentation of Diverse Qualitative Results}
\begin{figure}[t]
    \centering
    \includegraphics[width=0.95\linewidth]{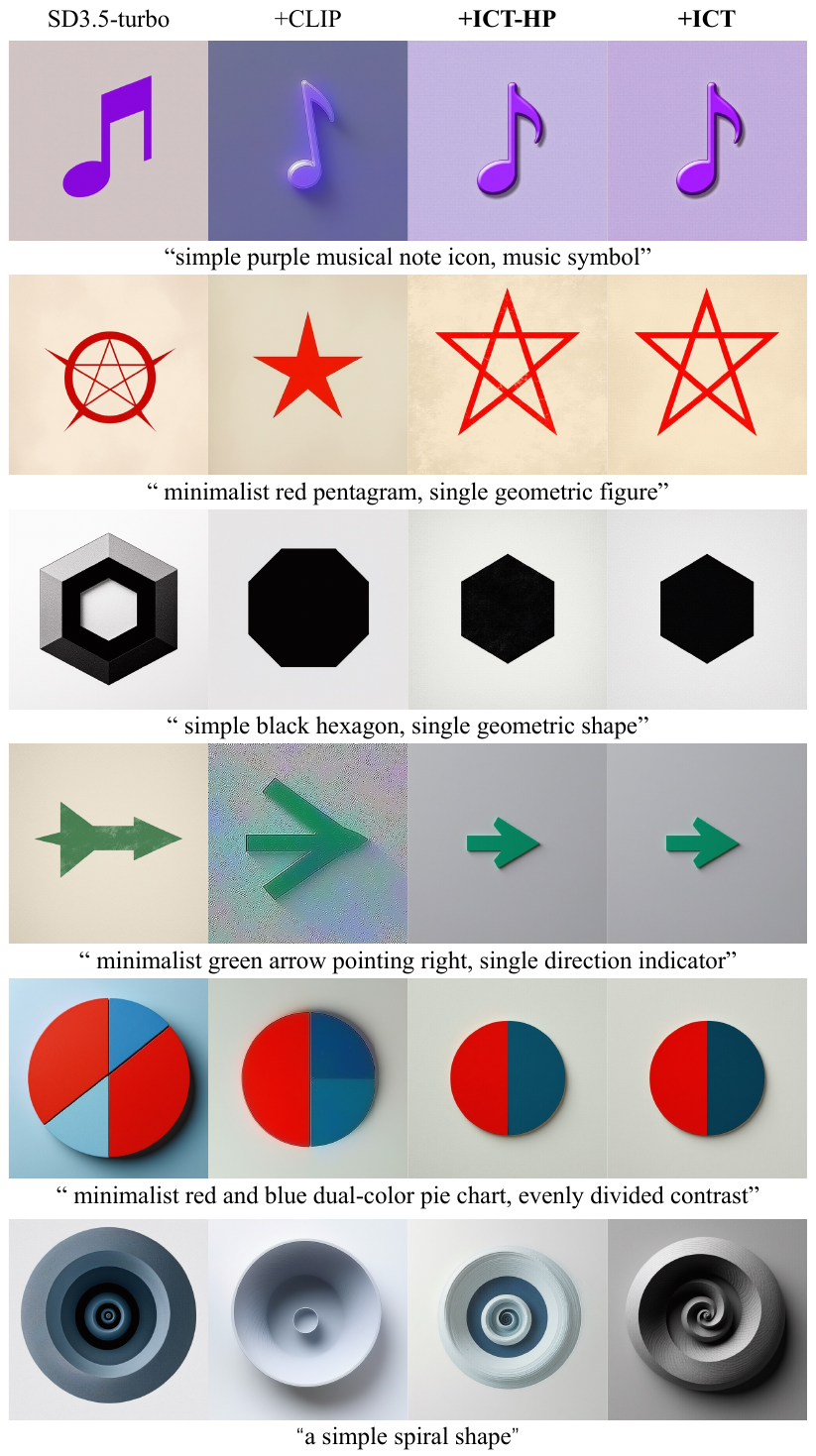}
    \caption{More simple elements generate results.}
    \label{fig:simple4}
\end{figure}

\noindent\textbf{Simple Element Generation Qualitative Results.}
To evaluate our model's performance in text-image alignment, particularly under minimal prompt conditions, we provide additional qualitative results of simple element generation, thoroughly demonstrating our method's excellence in text-image consistency. As shown in Figure \ref{fig:simple4}, we compared the original SD3.5-turbo with results optimized by CLIP model, ICT model, and ICT-HP model. Through detailed observation, it is evident that the original SD3.5-turbo exhibits significant limitations when executing minimal instructions; while the CLIP model-optimized version improves text-image alignment in some scenarios, it performs inconsistently across various situations. Among all variants, the ICT model optimization solution demonstrates the most precise and efficient effects, perfectly meeting all prompt requirements; meanwhile, the ICT-HP model optimization solution achieves near-optimal performance, with overall quality significantly superior to the CLIP model-optimized version. These experimental results strongly confirm that our innovative method not only significantly enhances the model's overall performance, but also successfully maintains and strengthens the base model's accurate understanding of clear, minimal concepts.

\begin{figure*}
    \centering
    \includegraphics[width=0.85\linewidth]{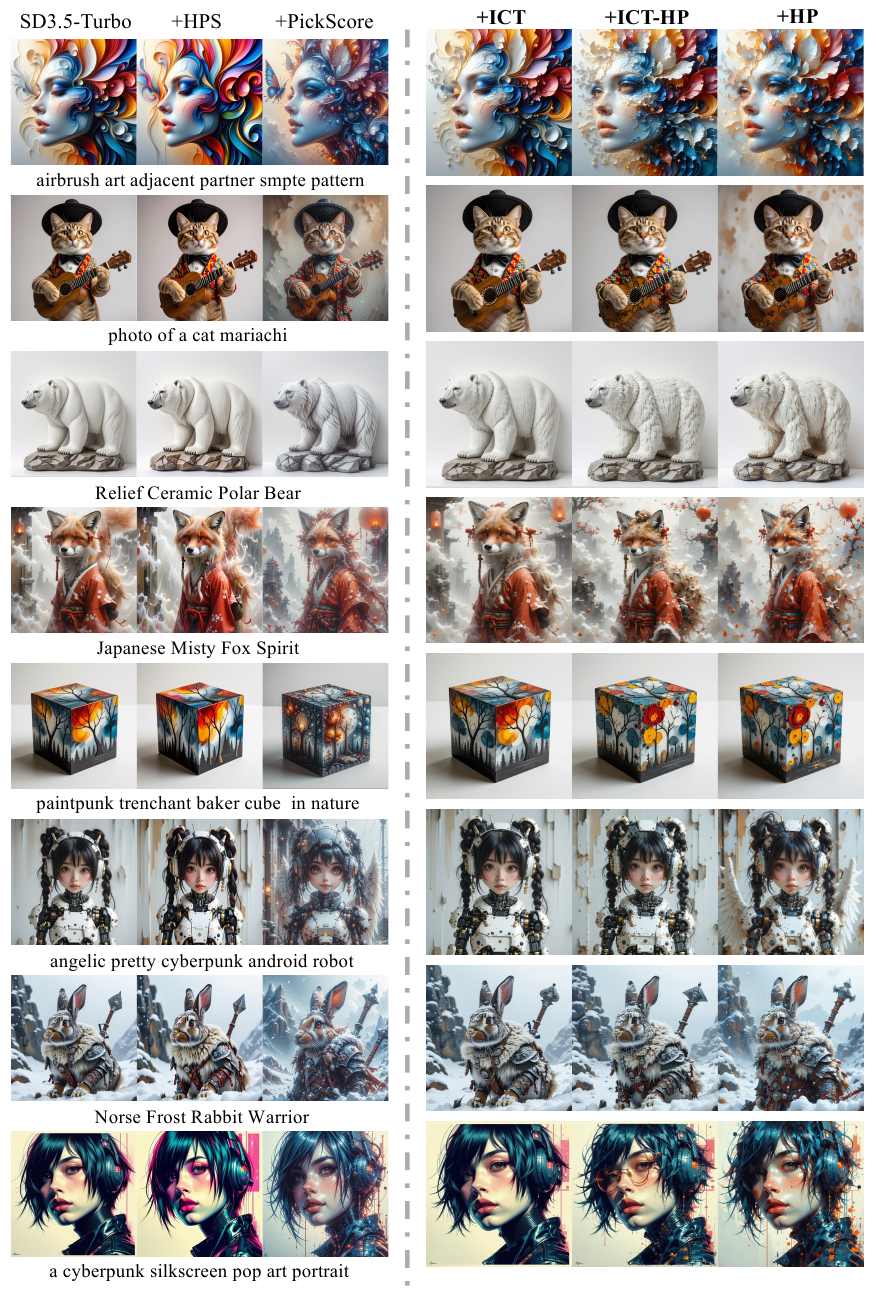}
    \caption{Qualitative Comparison of Optimization Results Across Reward Models Using Real User Prompts.}
    \label{fig:dx2}
\end{figure*}

\noindent\textbf{Qualitative Comparison of Optimization Results.}
In Figure \ref{fig:dx2}, we provide more examples of various reward optimizations. Based on our observations, ImageReward struggles to further optimize the high-performance diffusion model SD3.5-Large-Turbo, therefore we have excluded the qualitative results of ImageReward in this instance.

\begin{figure*}
    \centering
    \includegraphics[width=0.85\linewidth]{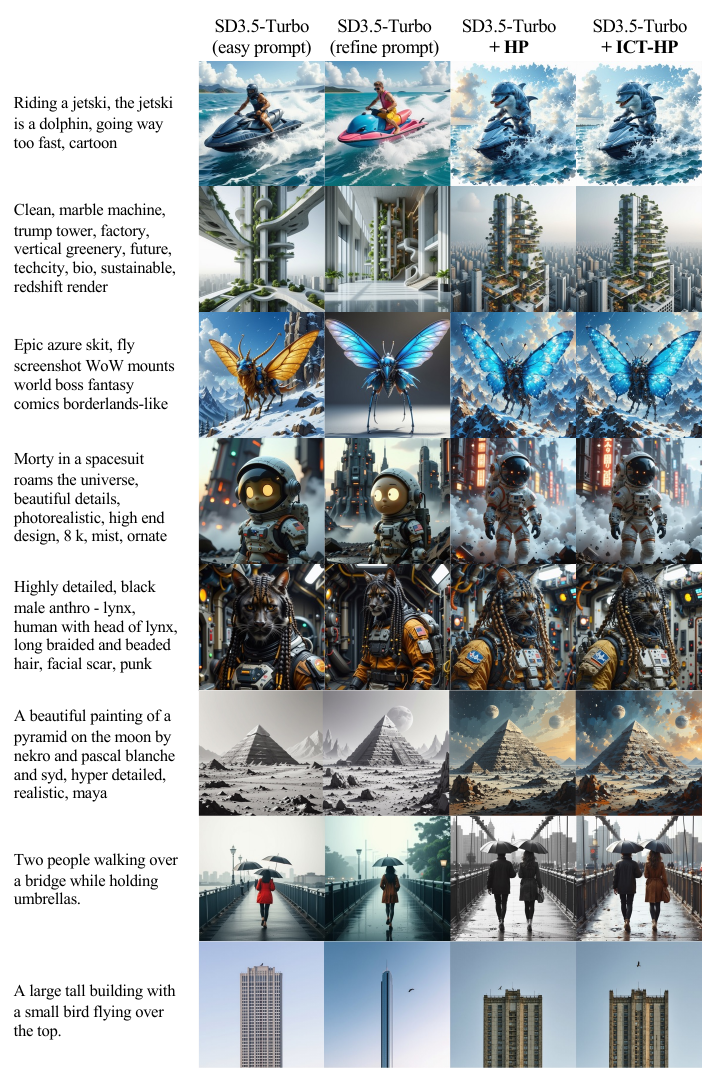}
    \caption{Qualitative Comparison: Origin Images, Refine Images, and Generations Optimized by Our Reward Models. }
    \label{fig:newtt2}
\end{figure*}

\noindent\textbf{Qualitative Comparison Between Optimization Results and Refine images.}
In Figure \ref{fig:newtt2}, we present comparative displays of additional original images, refine images, and results optimized through our model, confirming that our reward model can surpass the performance of prompt refinement.

\begin{figure*}
    \centering
    \begin{subfigure}[b]{1\linewidth}
        \centering
        \includegraphics[width=0.9\linewidth]{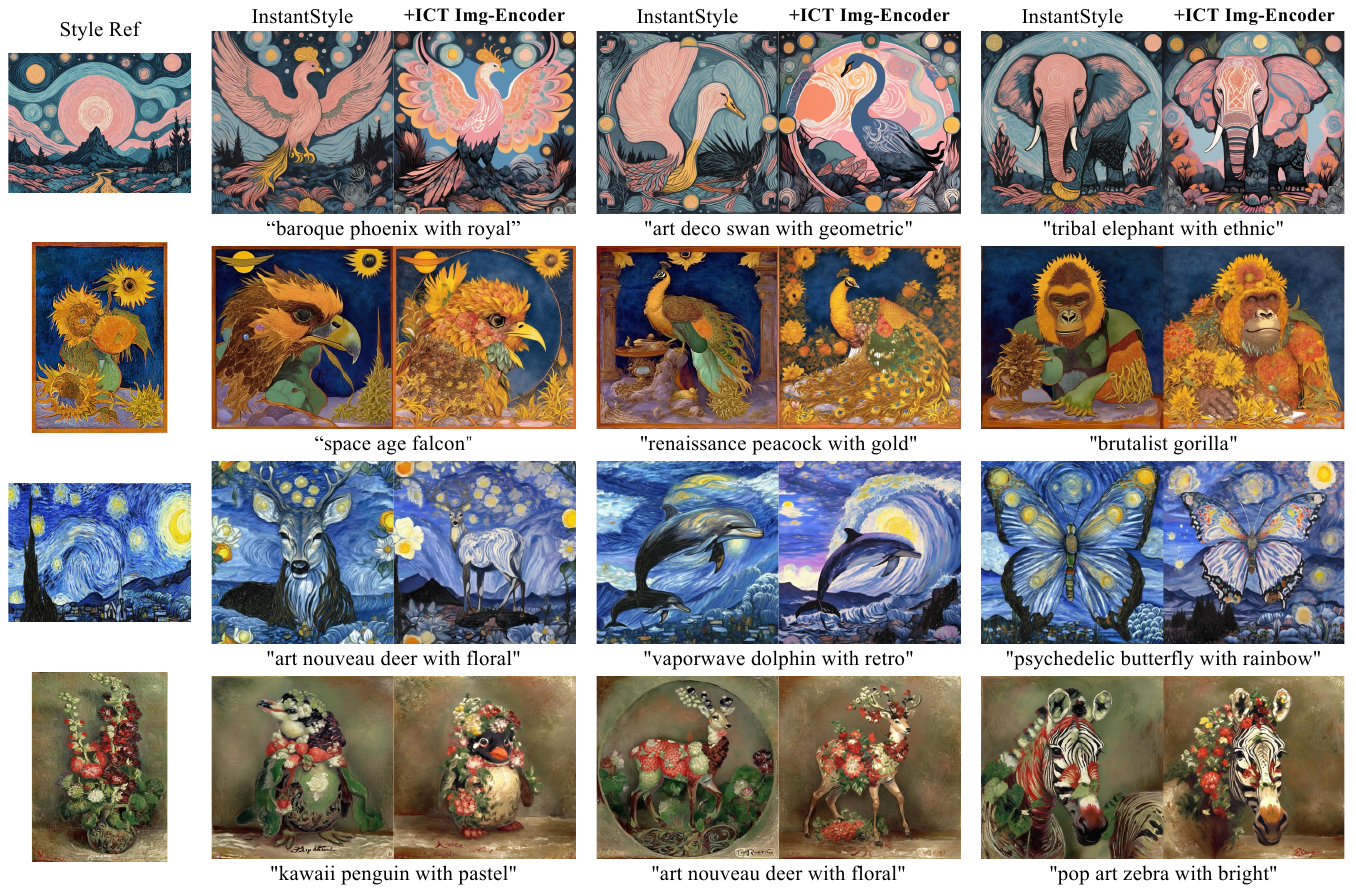}
        \caption{Qualitative Comparison of ICT Image-Encoder Performance in Style Injection Tasks.}
        \label{fig:style12}
    \end{subfigure}
    
    \vspace{0.3cm}
    
    \begin{subfigure}[b]{1\linewidth}
        \centering
        \includegraphics[width=0.9\linewidth]{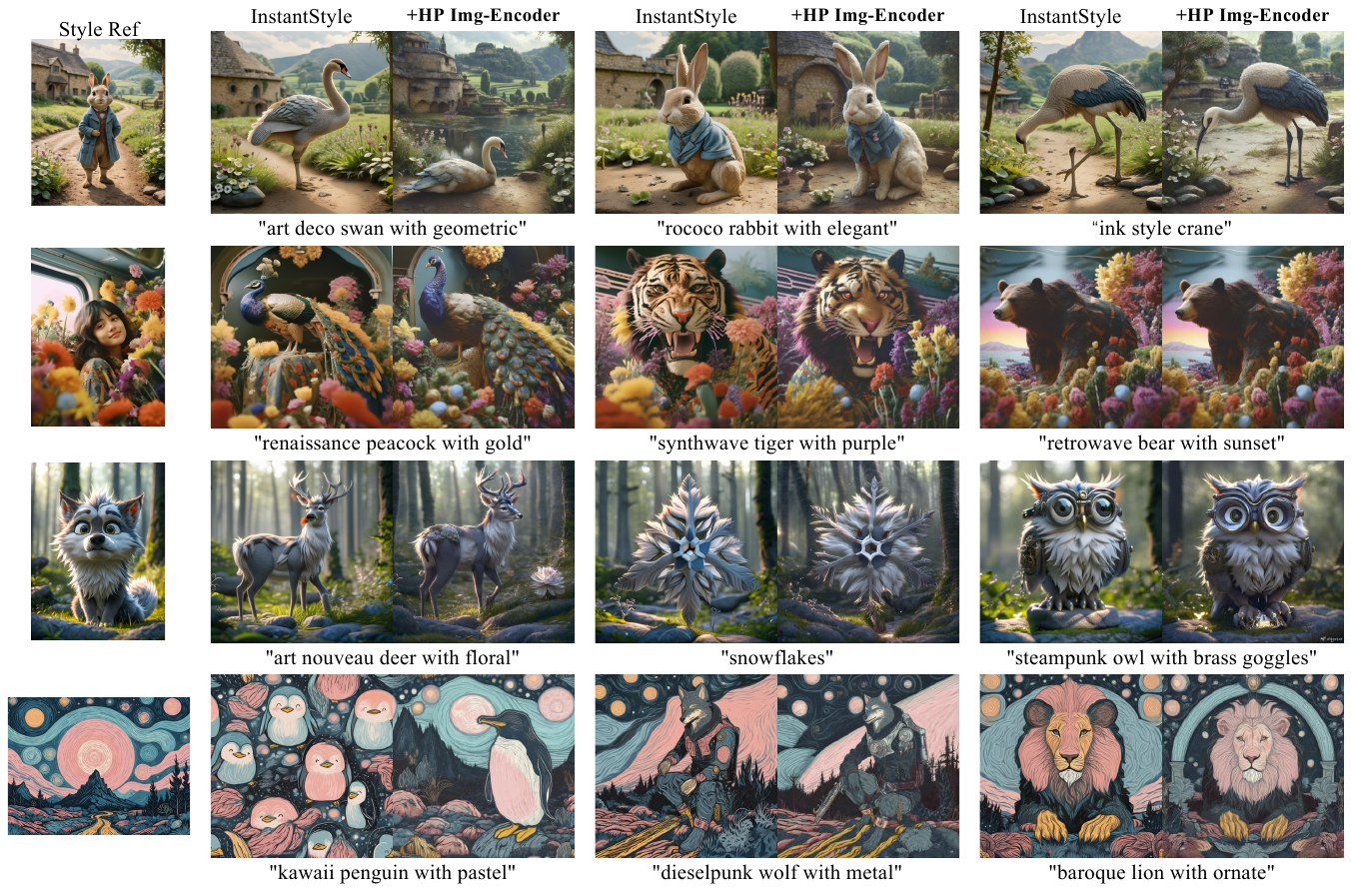}
        \caption{Qualitative Comparison of HP Image-Encoder Performance in Style Injection Tasks.}
        \label{fig:newtt}
    \end{subfigure}
    \caption{Qualitative Results of Style Injection Tasks.}
    \label{fig:combined}
\end{figure*}

\noindent\textbf{Qualitative Comparison of Style Injection Results.}
In Figure \ref{fig:combined}, we showcase additional qualitative results demonstrating style transfer achieved by extracting stylistic elements using Image-Encoders from both the HP model and the ICT model.

\section{Mathematical Framework and Theoretical Foundation of ICT-HP Reward System}

\subsection{Information Saturation Hypothesis and ICT Metric Formulation}

We propose the Information Saturation Hypothesis as mathematical foundation of ICT:

\begin{tcolorbox}[
 colback=gray!10,         
 colframe=gray!40,       
 sharp corners,          
 boxrule=0.5pt,          
 left=6pt,               
 right=6pt,              
 top=6pt,                
 bottom=6pt,             
 toptitle=3pt,           
 bottomtitle=3pt,        
 fonttitle=\bfseries\footnotesize, 
]
\footnotesize
\begin{hypothesis}[\textit{\textbf{Information Saturation Hypothesis}}]
\textit{ \\
For any image-text pair $(v,t)$, there exists a mutual information critical value $I^*(v,t)$ such that $v$ semantically aligns with $t$ if and only if $I(v;t) \geq I^*(v,t)$.
}
\end{hypothesis}
\end{tcolorbox}
As detailed in Section~\ref{sec:theory}, the CLIP score is formulated as 
\begin{equation}
\text{CLIP}(v,t) \approx \frac{I(v;t)}{\sqrt{(I(v;t) + I(v|t)) \cdot I(t)}}
\end{equation}
When $I(v;t)=I^*(v,t)$, according to the Information Saturation Hypothesis, the image fully contains the textual information, and 
\begin{equation}
\text{CLIP}^*(v,t) \approx \frac{I^*(v,t)}{\sqrt{(I^*(v,t) + I(v|t)) \cdot I(t)}}
\end{equation}
Since this function decreases monotonically with $I(v|t)$, its minimum value occurs at the critical threshold when $I(v|t)=I_{\max}(v|t)$. To ensure all semantically aligned image-text pairs receive an ICT score of 1, we set 
\begin{equation}
\theta^* = \text{CLIP}^*_{\min} \approx \frac{I^*(v,t)}{\sqrt{(I^*(v,t) + I_{\max}(v|t)) \cdot I(t)}}
\end{equation}
Physically, ICT reflects boundary saturation effect in human perception, where once an image adequately represents text, further details don't reduce alignment score.

\subsection{Preference Modeling and HP Metric Derivation}

Our data contains image triplets $(I_1, I_2, I_3)$ from identical prompts with preference hierarchy $I_3 \succ I_2 \succ I_1$, justifying HP's image-only approach. Bradley-Terry models preferences as: 
\begin{equation}
P(I_j \succ I_i) = \frac{1}{1 + \exp(-(s(I_j) - s(I_i)))}
\end{equation}
The log-likelihood: 
\begin{equation}
\mathcal{L} = \log P(I_2 \succ I_1) + \log P(I_3 \succ I_2)
\end{equation}
has negative upper bound via convex function theory, yielding ranking loss of HP Metric: 
\begin{equation}
\begin{aligned}
L_{margin} = \sum &[\max(0, -\Delta(I_2, I_1) + m) \\
&+ \max(0, -\Delta(I_3, I_2) + m)]
\end{aligned}
\end{equation}

\subsection{Multiplicative Integration Theory and System Properties}

Based on our analysis of CLIP scoring limitations, we propose a multiplicative dual-metric evaluation system:
\begin{equation}
\text{Reward}(v,t) = \text{ICT}(v,t) \cdot \text{HP}(v),
\end{equation}
where ICT and HP are defined as:
\begin{equation}
\text{ICT}(v,t) = \frac{I(v;t)}{I(t)},
\end{equation}
\begin{equation}
\text{HP}(v) = f(I(v|t)).
\end{equation}

This multiplicative formulation offers significant theoretical advantages:

\begin{enumerate}
\item \textbf{Complementary Constraint}: ICT ensures faithful textual expression while HP evaluates aesthetic quality; their product requires simultaneous satisfaction of both conditions.

\item \textbf{Threshold Effect}: When either metric approaches zero, the overall reward approaches zero, preventing optimization of one aspect at the expense of the other.

\item \textbf{Non-linear Gain}: When both ICT and HP improve simultaneously, the reward function exhibits accelerated growth, incentivizing concurrent enhancement of text containment and image quality.
\end{enumerate}

In information-theoretic terms, the multiplicative form can be expressed as:
\begin{equation}
\text{Reward}(v,t) \approx \frac{I(v;t)}{I(t)} \cdot f(I(v|t)),
\end{equation}

Compared to CLIP scoring:
\begin{equation}
\text{CLIP}(v,t) \approx \frac{I(v;t)}{\sqrt{I(t) \cdot (I(v;t) + I(v|t))}}.
\end{equation}

Our multiplicative formulation avoids the negative impact of $I(v|t)$ on the denominator in CLIP, instead positively utilizing $I(v|t)$ through the HP term, enabling accurate assessment of high-quality images.

\begin{figure}[t]
    \centering
    \includegraphics[width=1\linewidth]{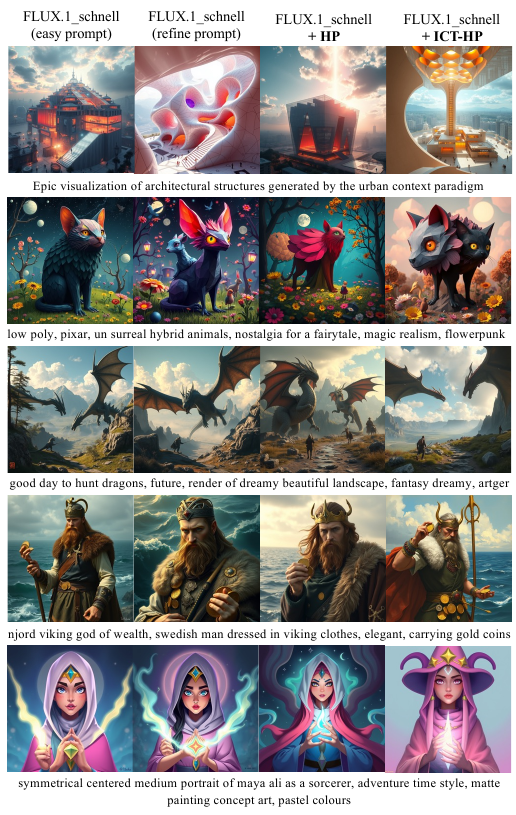}
    \caption{Qualitative Results of Optimizing FLUX.1-schnell.}
    \label{fig:flux1}
\end{figure}

\newpage

\newpage

\newpage

%% file: tabs/ladder.tex
\begin{table*}[t]
\centering
\setlength{\tabcolsep}{9pt}
\small
\begin{tabular}{l S[table-format=1.3] S[table-format=1.3] S[table-format=1.3] S[table-format=1.3] S[table-format=2.3] S[table-format=1.3] S[table-format=1.3] S[table-format=1.3] S[table-format=1.3]}
\toprule
\textbf{Model} & {\textbf{CLIP}$\boldsymbol{\uparrow}$} & {\textbf{ITM}$\boldsymbol{\uparrow}$} & {\textbf{ImgRwd}$\boldsymbol{\uparrow}$} & {\textbf{HPS}$\boldsymbol{\uparrow}$} & {\textbf{Pick}$\boldsymbol{\uparrow}$} & {\textbf{Aes}$\boldsymbol{\uparrow}$} & {\textbf{ICT}$\boldsymbol{\uparrow}$} & {\textbf{HP}$\boldsymbol{\uparrow}$} & {\textbf{ICT-HP}$\boldsymbol{\uparrow}$} \\
\midrule
SD1.5$_{e}$ & 0.336 & 0.697 & 0.188 & 0.267 & 20.689 & 5.884 & 0.667 & 0.672 & 0.457 \\
SD1.5$_{r}$ & 0.309 \textcolor{red}{$\boldsymbol{\downarrow}$} & 0.575 \textcolor{red}{$\boldsymbol{\downarrow}$} & -0.078 \textcolor{red}{$\boldsymbol{\downarrow}$} & 0.264 \textcolor{red}{$\boldsymbol{\downarrow}$} & 20.395 \textcolor{red}{$\boldsymbol{\downarrow}$} & 6.169 \textcolor{green}{$\boldsymbol{\uparrow}$} & 0.680 \textcolor{green}{$\boldsymbol{\uparrow}$} & 0.685 \textcolor{green}{$\boldsymbol{\uparrow}$} & 0.477 \textcolor{green}{$\boldsymbol{\uparrow}$} \\
\midrule
SDXL$_{e}$ & 0.367 & 0.807 & 0.719 & 0.272 & 21.853 & 6.446 & 0.898 & 0.772 & 0.695 \\
SDXL$_{r}$ & 0.338 \textcolor{red}{$\boldsymbol{\downarrow}$} & 0.715 \textcolor{red}{$\boldsymbol{\downarrow}$} & 0.687 \textcolor{red}{$\boldsymbol{\downarrow}$} & 0.271 \textcolor{red}{$\boldsymbol{\downarrow}$} & 21.789 \textcolor{red}{$\boldsymbol{\downarrow}$} & 6.857 \textcolor{green}{$\boldsymbol{\uparrow}$} & 0.908 \textcolor{green}{$\boldsymbol{\uparrow}$} & 0.776 \textcolor{green}{$\boldsymbol{\uparrow}$} & 0.705 \textcolor{green}{$\boldsymbol{\uparrow}$} \\
\midrule
SD3.5-T$_{e}$ & 0.340 & 0.831 & 0.955 & 0.277 & 22.034 & 6.555 & 0.910 & 0.777 & 0.717 \\
SD3.5-T$_{r}$ & 0.326 \textcolor{red}{$\boldsymbol{\downarrow}$} & 0.757 \textcolor{red}{$\boldsymbol{\downarrow}$} & 0.914 \textcolor{red}{$\boldsymbol{\downarrow}$} & 0.276 \textcolor{red}{$\boldsymbol{\downarrow}$} & 22.012 \textcolor{red}{$\boldsymbol{\downarrow}$} & 6.780 \textcolor{green}{$\boldsymbol{\uparrow}$} & 0.923 \textcolor{green}{$\boldsymbol{\uparrow}$} & 0.781 \textcolor{green}{$\boldsymbol{\uparrow}$} & 0.718 \textcolor{green}{$\boldsymbol{\uparrow}$} \\
\midrule
SD3.5-L$_{e}$ & 0.356 & 0.881 & 1.038 & 0.277 & 21.950 & 6.403 & 0.943 & 0.777 & 0.732 \\
SD3.5-L$_{r}$ & 0.334 \textcolor{red}{$\boldsymbol{\downarrow}$} & 0.800 \textcolor{red}{$\boldsymbol{\downarrow}$} & 0.955 \textcolor{red}{$\boldsymbol{\downarrow}$} & 0.276 \textcolor{red}{$\boldsymbol{\downarrow}$} & 21.853 \textcolor{red}{$\boldsymbol{\downarrow}$} & 6.704 \textcolor{green}{$\boldsymbol{\uparrow}$} & 0.939\textcolor{red} 
{$\boldsymbol{\downarrow}$}  & 0.779 \textcolor{green}{$\boldsymbol{\uparrow}$} & 0.731 \textcolor{red}{$\boldsymbol{\downarrow}$} \\
\midrule
FLUX\_S$_{e}$ & 0.349 & 0.866 & 0.974 & 0.277 & 21.739 & 6.439 & 0.891 & 0.763 & 0.683 \\
FLUX\_S$_{r}$ & 0.335 \textcolor{red}{$\boldsymbol{\downarrow}$} & 0.800 \textcolor{red}{$\boldsymbol{\downarrow}$} & 0.893 \textcolor{red}{$\boldsymbol{\downarrow}$} & 0.276 \textcolor{red}{$\boldsymbol{\downarrow}$} & 21.657 \textcolor{red}{$\boldsymbol{\downarrow}$} & 6.691 \textcolor{green}{$\boldsymbol{\uparrow}$} & 0.909 \textcolor{green}{$\boldsymbol{\uparrow}$} & 0.773 \textcolor{green}{$\boldsymbol{\uparrow}$} & 0.704 \textcolor{green}{$\boldsymbol{\uparrow}$} \\
\midrule
FLUX$_{e}$ & 0.333 & 0.818 & 0.982 & 0.279 & 22.021 & 6.642 & 0.906 & 0.775 & 0.703 \\
FLUX$_{r}$ & 0.326 \textcolor{red}{$\boldsymbol{\downarrow}$} & 0.791 \textcolor{red}{$\boldsymbol{\downarrow}$} & 0.967 \textcolor{red}{$\boldsymbol{\downarrow}$} & 0.278 \textcolor{red}{$\boldsymbol{\downarrow}$} & 21.899 \textcolor{red}{$\boldsymbol{\downarrow}$} & 6.846 \textcolor{green}{$\boldsymbol{\uparrow}$} & 0.919 \textcolor{green}{$\boldsymbol{\uparrow}$} & 0.778 \textcolor{green}{$\boldsymbol{\uparrow}$} & 0.715 \textcolor{green}{$\boldsymbol{\uparrow}$} \\
\bottomrule
\end{tabular}
\caption{\textbf{Quantitative Results of Image Generation Models}.}
\label{tab:ladder}
\end{table*}

%% file: tabs/geneval_fl.tex
\begin{table*}
\centering
\setlength{\tabcolsep}{2pt}
\footnotesize
\resizebox{\textwidth}{!}{ 
\begin{tabular}{@{}lccccccc@{}}
\toprule
\textbf{Model} &
{\bf Mean $\uparrow$ } &
{\bf Single$\uparrow$} &
{\bf Two$\uparrow$} &
{\bf Counting$\uparrow$} &
{\bf Colors$\uparrow$} &
{\bf Position$\uparrow$} &
{\bf Color Attribution$\uparrow$}\\
\midrule
SDXL & 0.55 & 0.98 & 0.74 & 0.39 & 0.85 & 0.15 & 0.23  \\
DALL-E 2 & 0.52 & 0.94 & 0.66 & 0.49 & 0.77 & 0.10 & 0.19  \\
DALL-E 3 & 0.67 & 0.96 & 0.87 & 0.47 & 0.83 & \textbf{0.43} & 0.45 \\
SD3 & 0.68 & 0.98 & 0.84 & 0.66 & 0.74 & \underline{0.40} & 0.43  \\
\midrule
FLUX.1-Schnell & 0.68 &  0.99 & 0.88 & 0.62 & 0.76 & 0.30 & 0.51 \\
\rowcolor[HTML]{E6E6FA} \textbf{+ ICT-HP (Ours)} & \textbf{0.69} & \textbf{0.99} & \textbf{0.88} & \textbf{0.62} & \textbf{0.81} & 0.29  & \textbf{0.81} \\
\rowcolor[HTML]{E6E6FA} \textbf{+ ICT (Ours)} & \textbf{0.69} & \textbf{0.99} & \textbf{0.88} & 0.60 & \textbf{0.81} & 0.29 & 0.55 \\
\midrule
FLUX.1-dev & 0.66 &  0.97 & 0.82 & 0.71 & 0.78 & 0.22 & 0.45 \\
\rowcolor[HTML]{E6E6FA} \textbf{+ ICT-HP (Ours)} & \textbf{0.67} & \textbf{0.99} & \textbf{0.82} & \textbf{0.74} & \textbf{0.80} & 0.20  & \textbf{0.51} \\
\rowcolor[HTML]{E6E6FA} \textbf{+ ICT (Ours)} & \textbf{0.67} & 0.98 & 0.81 & 0.73 & \textbf{0.80} & 0.19 & 0.50 \\
\bottomrule
\end{tabular}
}
\caption{\textbf{Quantitative GenEval Results of FLUX.1-schnell with RM-Optimized LoRA (HP/ICT-HP) and Transfer Performance on FLUX.1-dev.}} 
\label{tab:geneval_fl}
\vspace{-1.5em}
\end{table*}